%% file: main.tex
\documentclass[journal]{IEEEtran}
\usepackage{amsmath,amsfonts}
\usepackage{algorithm}
\usepackage{array}
\usepackage{textcomp}
\usepackage{stfloats}
\usepackage{url}
\usepackage{verbatim}
\usepackage{graphicx}
\usepackage{cite}
\usepackage{amsmath,amssymb,amsfonts,amsthm,bm}
\usepackage{algorithmicx}
\usepackage{algpseudocode}
\usepackage{support-caption}
\usepackage{subcaption}
\usepackage{graphicx}
\usepackage{longtable}
\usepackage{multirow}
\usepackage{tabularx}
\usepackage{booktabs}
\usepackage{textcomp}
\usepackage{stfloats}
\usepackage{url}
\usepackage{verbatim}
\usepackage{graphicx}
\usepackage{color}
\usepackage[acronyms]{glossaries}

\algnewcommand\algorithmicinput{\textbf{Input:}}
\algnewcommand\algorithmicoutput{\textbf{Output:}}
\algnewcommand\INPUT{\item[\algorithmicinput]} 
\algnewcommand\OUTPUT{\item[\algorithmicoutput]}

\usepackage[hidelinks]{hyperref}

\usepackage{enumitem}

\newcommand{\ie}{{\emph{i.e.}}}
\newcommand{\etal}{{\emph{et al.}}}
\newcommand{\eg}{{\emph{e.g.}}}
\newcommand{\wrt}{{\emph{w.r.t.}}}

\newacronym{sbl}{SBL}{sparse Bayesian learning}
\newacronym{ml}{ML}{maximum likelihood}
\newacronym{pdf}{PDF}{probability density function}
\newacronym{snr}{SNR}{signal-to-noise ratio}
\newacronym{em}{EM}{expectation-maximization}
\newacronym{awgn}{AWGN}{additive white Gaussian noise}
\newacronym{nmse}{NMSE}{normalized mean squared error}
\newacronym{mmse}{MMSE} {minimum mean squared error}
\newacronym{elbo}{ELBO}{evidence lower bound}
\newacronym{kl}{KL}{Kullback-Leibler}

\begin{document}

\title{Enhancing Gradient Inversion Attacks in Federated Learning via Hierarchical Feature Optimization}

    \author{Hao Fang$^*$,
    Wenbo Yu$^*$,
    Bin Chen$^{\dagger}$,~\IEEEmembership{Member,~IEEE,}
    Xuan Wang,~\IEEEmembership{Member,~IEEE,}\\
    Shu-Tao Xia,~\IEEEmembership{ Member,~IEEE,}
    Qing Liao,~\IEEEmembership{Member,~IEEE,}
    Ke Xu,~\IEEEmembership{Fellow,~IEEE}
    \IEEEcompsocthanksitem
    \IEEEcompsocitemizethanks{
    \IEEEcompsocthanksitem Hao Fang, Wenbo Yu, and Shu-Tao Xia are with the Tsinghua Shenzhen International Graduate School, Tsinghua University, Shenzhen 518055, Guangdong, China (e-mail: \{ffhibnese, wenbo.research\}@gmail.com; xiast@sz.tsinghua.edu.cn).
    \IEEEcompsocthanksitem Bin Chen, Xuan Wang, and Qing Liao are with the School of Computer Science and Technology, Harbin Institute of Technology, Shenzhen 518055, Guangdong, China (e-mail: chenbin2021@hit.edu.cn; wangxuan@cs.hitsz.edu.cn, liaoqing@hit.edu.cn). 
    \IEEEcompsocthanksitem Ke Xu is with the Department of Computer Science and Technology, Tsinghua University, Beijing 100084, China (e-mail: xuke@tsinghua.edu.cn).
    \IEEEcompsocthanksitem $^*$Hao Fang and Wenbo Yu contribute equally to this paper.
    \IEEEcompsocthanksitem $^{\dagger}$Corresponding author: Bin Chen (e-mail: chenbin2021@hit.edu.cn).
    }
}



\maketitle

\begin{abstract}
Federated Learning (FL) has emerged as a compelling paradigm for privacy-preserving distributed machine learning, allowing multiple clients to collaboratively train a global model by transmitting locally computed gradients to a central server without exposing their private data. Nonetheless, recent studies find that the gradients exchanged in the FL system are also vulnerable to privacy leakage, e.g., an attacker can invert shared gradients to reconstruct sensitive data by leveraging pre-trained generative adversarial networks (GAN) as prior knowledge. 
However, existing attacks simply perform gradient inversion in the latent space of the GAN model, which limits their expression ability and generalizability. To tackle these challenges, we propose \textbf{G}radient \textbf{I}nversion over \textbf{F}eature \textbf{D}omains (GIFD), which disassembles the GAN model and searches the hierarchical features of the intermediate layers. Instead of optimizing only over the initial latent code, we progressively change the optimized layer, from the initial latent space to intermediate layers closer to the output images. In addition, we design a regularizer to avoid unreal image generation by adding a small ${l_1}$ ball constraint to the searching range. We also extend GIFD to the out-of-distribution (OOD) setting, which weakens the assumption that the training sets of GANs and FL tasks obey the same data distribution. Furthermore, we consider the challenging OOD scenario of label inconsistency and propose a label mapping technique as an effective solution. Extensive experiments demonstrate that our method can achieve high-quality reconstruction and outperform competitive baselines across a variety of FL scenarios. 
\end{abstract}


\input{introduction}
\input{related_work}
\input{method}
\input{Experiments}

\input{conclusions}

\bibliographystyle{IEEEtran}
\bibliography{Bibliography}
\newpage
\section{Biography Section}
\vspace{-25pt}
\begin{IEEEbiography}[{\includegraphics[width=1in,height=1.25in,clip,keepaspectratio]{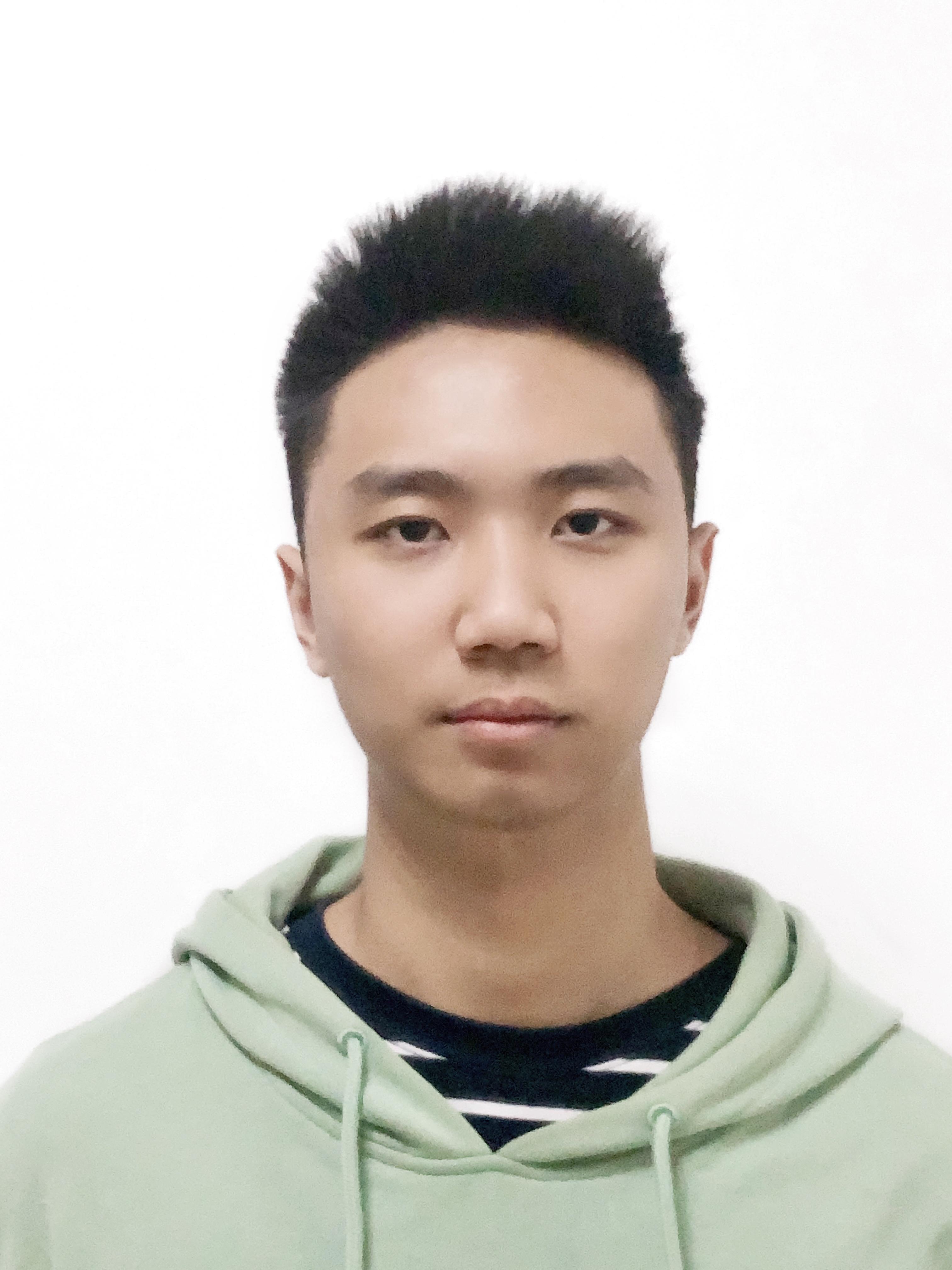}}]{Hao Fang} received a bachelor’s degree from the School of Computer Science and Technology, Harbin Institute of Technology, Shenzhen, China, in 2023. He is currently pursuing a Ph.D. degree in Computer Science and Technology from Tsinghua Shenzhen International School, Tsinghua University, China. His research interests include trustworthy AI and large language models, especially in MLLM/Agentic reinforcement learning. He has published in top-tier journals and conferences, such as IJCV, IEEE TIFS, CVPR/ICCV/ECCV, NeurIPS/ICLR, and ACL/EMNLP. He has also served as a reviewer for top-tier journals and conferences, including IEEE TPAMI, IEEE TSDC, IEEE TIFS, IEEE TMC, NeurIPS/ICLR/ICML, and CVPR/ECCV.
\end{IEEEbiography}
\vspace{-25pt}
\begin{IEEEbiography}[{\includegraphics[width=1in,height=1.25in,clip,keepaspectratio]{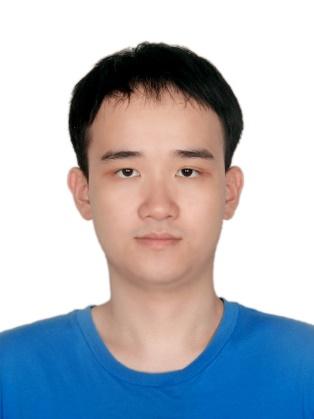}}]{Wenbo Yu}
received a bachelor's degree from the School of Computer Science and Technology, Harbin Institute of Technology, Shenzhen, China, in June 2024. He is currently pursuing a master's degree in Computer Technology at the Tsinghua Shenzhen International Graduate School, Tsinghua University, China. His research interests mainly include large language models and AI security. He has published articles in top-tier journals and conferences, such as IEEE TIFS and EMNLP. He has also been invited to serve as a reviewer for many top-tier journals and conferences, including IEEE JSAC and WWW.
\end{IEEEbiography}
\vspace{-25pt}
\begin{IEEEbiography}[{\includegraphics[width=1in,height=1.25in,clip,keepaspectratio]{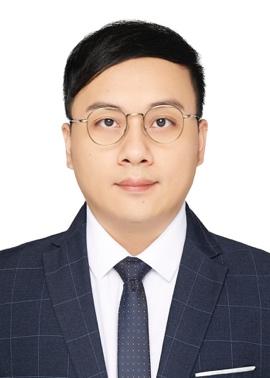}}]{Bin Chen}
(Member, IEEE) received the B.S. and M.S. degrees in mathematics from South China Normal University, Guangzhou, China, in 2014 and 2017, respectively, and the Ph.D. degree in computer science and technology from the Department of Computer Science and Technology, Tsinghua University, Beijing, China, in 2021. From December 2019 to May 2020, he visited the Department of Electrical and Computer Engineering, University of Waterloo, Waterloo, ON, Canada. From May 2021 to November 2021, he was a Post-Doctoral Researcher with Shenzhen International Graduate School, Tsinghua University, Shenzhen, China. Since December 2021, he has been a Professor with the Department of Computer Science and Technology, Harbin Institute of Technology (Shenzhen). He has published more than 100 technical papers at prominent journals and conferences, such as IEEE Trans. on Pattern Analysis and Machine Intelligence, IEEE Trans. on Multimedia, International Journal of Computer Vision, ICML, NeurIPS, ICLR, CVPR, ICCV, ECCV, AAAI, and ACM MM. His research interests include information theory, multimedia compression, and AI safety. He served as the Associate Editor of IEEE Trans. on Multimedia, and served as the Area Chair for NeurIPS, ICLR, and AAAI.
\end{IEEEbiography}
\vspace{-25pt}
\begin{IEEEbiography}[{\includegraphics[width=1in,height=1.25in,clip,keepaspectratio]{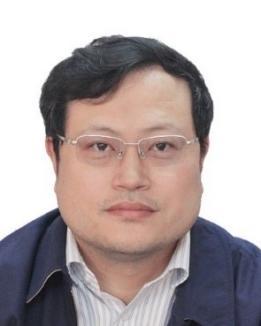}}]{Xuan Wang}
(Member, IEEE) received the Ph.D. degree in computer science from Harbin Institute of Technology in 1997. He is one of the inventors of Microsoft Pinyin, and once worked in Microsoft headquarter in Seattle due to his contribution to Microsoft Pinyin. He is currently a professor of the School of Computer Science and Technology, Harbin Institute of Technology, Shenzhen, China. His main research interests include cybersecurity, information game theory, and artificial intelligence.
\end{IEEEbiography}
\vspace{-25pt}
\begin{IEEEbiography}[{\includegraphics[width=1in,height=1.25in,clip,keepaspectratio]{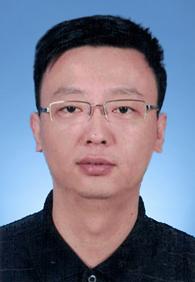}}]{Shu-Tao Xia}
(Member, IEEE) received the B.S. degree in mathematics and the Ph.D. degree in applied mathematics from Nankai University, Tianjin, China, in 1992 and 1997, respectively. From March 1997 to April 1999, he was with the Research Group of Information Theory, Department of Mathematics, Nankai University. Since January 2004, he has been with Tsinghua Shenzhen International Graduate School, Tsinghua University, Guangdong, China, where he is currently a Full Professor. His papers have been published in multiple top-tier journals and conferences, such as IEEE TPAMI, IEEE TIFS, IEEE TDSC, CVPR, ICLR, ICCV, and NeurIPS. His current research interests include coding and information theory, networking, machine learning, and AI security.
\end{IEEEbiography}
\vspace{-25pt}
\begin{IEEEbiography}[{\includegraphics[width=1in,height=1.25in,clip,keepaspectratio]{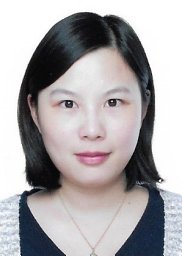}}]{Qing Liao} received the Ph.D. degree from Hong Kong University of Science and Technology in 2016. She is currently a professor with School of Computer Science and Technology, Harbin Institute of Technology (Shenzhen), China. Her research interests include data mining, artificial intelligence and information security, etc. She is a member of IEEE since 2013.
\end{IEEEbiography}
\vspace{-20pt}
\begin{IEEEbiography}[{\includegraphics[width=1in,height=1.25in,clip,keepaspectratio]{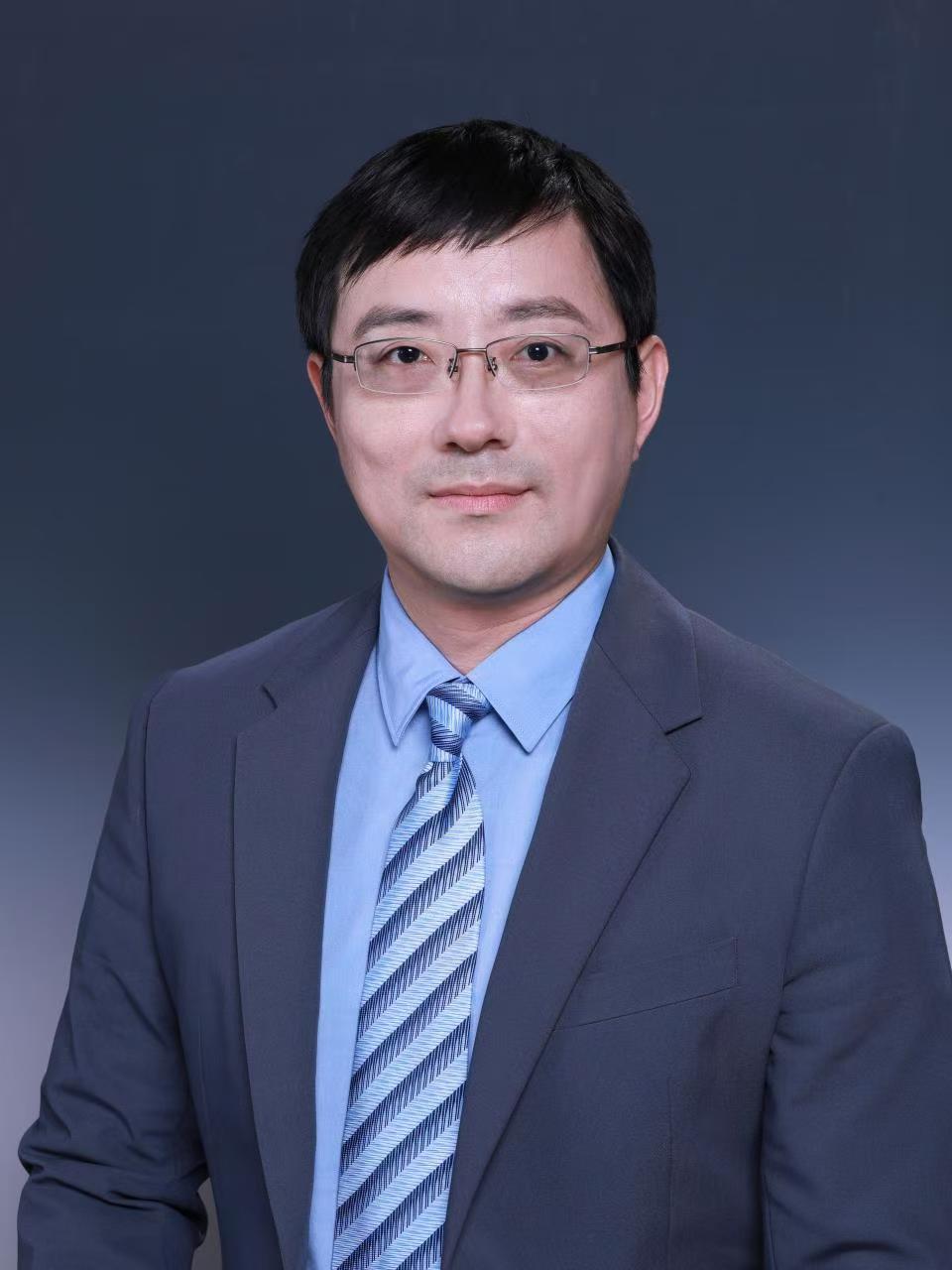}}]{Ke Xu}
(Fellow, IEEE) received the Ph.D. degree from the Department of Computer Science and Technology, Tsinghua University, Beijing, China. He is currently a Full Professor in the Department of Computer Science and Technology, Tsinghua University. He has published more than 200 technical articles and holds 11 U.S. patents in the research areas of next-generation Internet, blockchain systems, the Internet of Things, and network security. He is a member of ACM and an IEEE Fellow. He was the Steering Committee Chair of IEEE/ACM IWQoS. He has guest-edited several special issues in IEEE and Springer journals.
\end{IEEEbiography}



\vfill

\end{document}

%% file: introduction.tex
\section{Introduction}
\IEEEPARstart{F}{ederated} learning \cite{mcmahan2017communication, zhang2021survey, yu2025gi} has become an increasingly popular distributed machine learning framework, which has been widely applied in many privacy-sensitive scenarios \cite{li2020review, yang2020federated, wen2023survey}, such as financial services, medical analysis, and recommendation systems. 
It allows multiple clients to collaboratively train a shared model under the coordination of the central server, which aggregates the uploaded gradients calculated from the local data by the end users, rather than exposing raw private data. This mechanism resolves the problem of data silos and brings privacy benefits to distributed learning. 
However, a series of recent studies have shown that even gradients uploaded in FL are at risk of privacy leakage. 
Specifically, Zhu \etal \cite{zhu2019deep} propose the first gradient leakage attack (GLA) and formulate the data recovery as an optimization problem of gradient matching. Given the gradient $g$ corresponding to a private batch of images $\mathbf{x^*}$ and labels $\mathbf{y^*}$ in FL, an attacker first initializes a local dummy batch $(\mathbf{\hat{x}}, \mathbf{\hat{y}})$ and computes its gradient $\hat{g}$. The attacker then minimizes the distance between $g$ and $\hat{g}$ to iteratively update $(\mathbf{\hat{x}}, \mathbf{\hat{y}})$, driving them towards the ground-truth (GT) data $(\mathbf{x^*}, \mathbf{y^*})$. Subsequent optimization-based methods further enhance attack effectiveness through GT label inference \cite{zhao2020idlg}, novel loss designs with image smoothness regularization \cite{geiping2020inverting}, and batch-level label extraction combined with feature distribution regularization based on batch normalization (BN) statistics \cite{yin2021see}.


\begin{figure}
    \begin{subfigure}{0.999\linewidth}
        \begin{minipage}[t]{0.05\linewidth}
        \rotatebox{90}{{\textbf{~ImageNet}}}   
        \end{minipage}%
        \begin{minipage}[t]{0.23\linewidth}
        \centering
        \includegraphics[width=1.7cm]{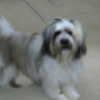}
        \centering

        \end{minipage}%
        \begin{minipage}[t]{0.23\linewidth}
        \centering
        \includegraphics[width=1.7cm]{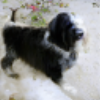}
        \centering

        \end{minipage}%
        \begin{minipage}[t]{0.23\linewidth}
        \centering
        \includegraphics[width=1.7cm]{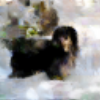}
        \centering

        \end{minipage}%
        \begin{minipage}[t]{0.23\linewidth}
        \centering
        \includegraphics[width=1.7cm]{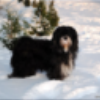}
        \centering

        \end{minipage}%
        \end{subfigure}
        
    \vskip 8pt
    
    \begin{subfigure}{1\linewidth}
        \begin{minipage}[t]{0.05\linewidth}
        \rotatebox{90}{{\textbf{~~~FFHQ}}}   
        \end{minipage}%
        \begin{minipage}[t]{0.23\linewidth}
        \centering
        \includegraphics[width=1.7cm]{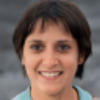}
        \centering
        \caption*{\textbf{\footnotesize{Dummy Input}}}
        \end{minipage}%
        \begin{minipage}[t]{0.23\linewidth}
        \centering
        \includegraphics[width=1.7cm]{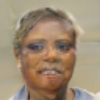}
        \centering
        \caption*{\textbf{\footnotesize{{Latent Space}}}}
        \end{minipage}%
        \begin{minipage}[t]{0.23\linewidth}
        \centering
        \includegraphics[width=1.7cm]{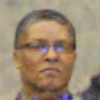}
        \centering
        \caption*{\textbf{\footnotesize{GIFD}}}
        \end{minipage}%
        \begin{minipage}[t]{0.23\linewidth}
        \centering
        \includegraphics[width=1.7cm]{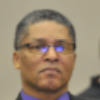}
        \centering
        \caption*{\textbf{\footnotesize{Ground Truth}}}
        \end{minipage}%
        \end{subfigure}

\caption{The reconstructed results of our proposed GIFD on ImageNet\cite{deng2009imagenet} and FFHQ\cite{karras2019style}. The first column contains the randomly initialized images generated by GAN models. The next two columns show the reconstruction samples of the latent space search and our proposed GIFD.} 
\label{contribution}
\end{figure}

It is widely acknowledged that a pre-trained GAN model learned from a public dataset captures a wealth of prior knowledge \cite{fang2024clip, fang2025one}. Motivated by this, recent attacks \cite{yin2021see, jeon2021gradient, li2022auditing} propose to leverage the manifold of GAN models as image priors to provide a good approximation of the natural image space and significantly enhance the attacks. Despite the success, these attacks simply restrict their search range to the latent or parameter space of GAN models, failing to fully exploit the rich information encoded in generative priors and leading to inaccurate reconstruction in some cases. Moreover, most of these works rely on strong assumptions, such as access to GT labels \cite{jeon2021gradient}, BN statistics \cite{yin2021see}, or the private data distribution \cite{jeon2021gradient, li2022auditing}, which are actually impractical in real FL scenarios. As a result, recovering high-quality private data under more realistic setups remains a significant challenge for most existing methods.

In this paper, we advocate a simple and effective solution, Gradient Inversion over Feature Domain (GIFD), to address the limited expressiveness and poor generalizability of pre-trained GANs in gradient inversion attacks. Recently, it has been shown that rich semantic information is encoded in the intermediate features and the latent space of GANs \cite{bau2019gan, tewari2020pie, shen2020interpreting,daras2021intermediate}. Inspired by these works, we reformulate the GAN inversion as a novel intermediate layer optimization problem that minimizes the gradient matching loss by searching the intermediate features of the generative model. Specifically, the optimization begins in the latent space and proceeds by successively fine-tuning the intermediate layers of the generative model. 
During feature domain optimization, the image is generated using a subset of the generator, and the higher dimensionality of intermediate features induces a considerably larger solution space, which may discard information captured in earlier layers and lead to unrealistic image generation. 
To solve this problem, we iteratively project the optimizing features to a small ${\ell_1}$ ball centered at the initial vector induced by the previous layer. Finally, we select output images from the layer with the corresponding least gradient matching loss as the final results. The visual comparison in Figure \ref{contribution} clearly demonstrates the superiority of optimizing intermediate feature domains, validating that the proposed GIFD better exploits the rich knowledge encoded in the generative models.

Another issue unsolved in GAN-based gradient attacks is the feasibility of private data reconstruction under more rigorous and realistic settings. 
Existing GAN-based methods often assume that the private FL data distribution is accessible to the attacker \cite{yin2021see, jeon2021gradient, li2022auditing}, allowing them to use a GAN model trained on the same image dataset. We then examine GIFD's generality under scenarios of out-of-distribution (OOD) image recovery and consider a style-shift setting, where the feature distribution of private images differs substantially from that of the GAN's training set. The remarkable improvement of our method over baselines on various OOD settings demonstrates the excellent generalizability of the proposed method, rendering GIFD a practical approach to reconstruct unseen private data. An even more challenging issue is the \textit{label inconsistency}, in which the labels of private images are entirely misaligned with the conditional labels used by the GAN model. To this end, we propose a label mapping technique as a self-correction mechanism to mitigate inaccurate supervision and enhance label reliability, further improving the attack effectiveness.
Furthermore, we discuss attacks against FL systems with various widely used defense strategies, including gradient sparsification \cite{Strom2015, aji-heafield-2017-sparse}, gradient clipping \cite{geyer2017differentially}, differential privacy  \cite{geyer2017differentially}, Soteria (\ie, perturbing the data representations) \cite{sun2021soteria}, and ATSPrivacy \cite{gao2023automatic}. These frequently used privacy defense approaches have been confirmed to exhibit high resilience against existing attacks, yet our method still achieves effective data reconstruction.
Extensive experiments have demonstrated the effectiveness of the GIFD attack, without relying on impractical assumptions such as access to BN statistics or private data distribution.
In summary, our main contributions are as follows:
\begin{itemize}
  \item We propose GIFD, a novel gradient inversion attack that exploits pre-trained generative models as data priors to reconstruct private data by inverting gradients through successive optimization of the latent code and intermediate features of the generator under an ${\ell_1}$-ball constraint.
  \item We demonstrate that this optimization strategy can be utilized to reconstruct private OOD data with diverse styles, confirming the impressive generalization ability of the proposed GIFD in more realistic settings without relying on strong or impractical assumptions.
  \item We conduct a systematic evaluation of the proposed method against state-of-the-art baselines across a wide range of scenarios, demonstrating that GIFD achieves consistent improvements over previous methods.
\end{itemize}

This paper extends our preliminary conference version \cite{fang2023gifd} with substantial new contributions in methodology, evaluation, and analysis. Specifically, the current work incorporates:
(1) a novel self-correction technique for label mapping to enhance reconstruction performance under challenging OOD scenarios with label inconsistency;
(2) a comprehensive evaluation framework for FL systems under GLA, considering practical factors including data heterogeneity and varying numbers of FL training rounds;
(3) a more systematic defense evaluation against advanced defense strategies;
(4) a comprehensive comparison with recent attacks such as diffusion-based methods;
(5) new comparisons across a wider range of datasets and global models, including diverse variants of Vision Transformer (ViT) \cite{han2022survey};
(6) detailed ablation studies that validate the effectiveness of each proposed component; and
(7) additional visualizations of reconstructed images and a comparison of the cost function over time, showing the balance of GIFD between effectiveness and efficiency compared to baselines.

%% file: related_work.tex
\section{RELATED WORK}
\subsection{Gradient-based Attack in FL}
Privacy attacks in FL have attracted sufficient attention \cite{egger2025federated, zhang2024functionality, ye2024feature, shi2025towards}. Early privacy studies investigate \emph{membership inference} \cite{shokri2017membership,melis2019exploiting} in federated learning, where a malicious attacker can determine whether a certain data sample has been involved in model training. A similar type of attack, called \emph{property inference} \cite{ganju2018property}, aims to reveal the attributes of private samples in the training set. Another powerful attack is \emph{model inversion} (MI) \cite{hitaj2017deep, fang2024privacy, qiu2024closer} that recovers distribution-level private data. Hitaj \etal \cite{hitaj2017deep} extended MI attacks to FL systems by manually training a GAN model from local images and the shared gradients to generate samples with the same distribution as the private data. Wang \etal \cite{wang2019beyond} then enhanced the attack efficacy and reconstructed client-level data representatives.

\textbf{Gradient Inversion Attacks.} 
Among different security threats \cite{fang2025language,fang2026retrievals, fang2025grounding,fang2026seeing,xu2026bypassing,xiao2026diffusion,xiao2026leveraging,yu2026chillguard}, gradient inversion represents a particularly threatening privacy attack, where an adversary can reconstruct a client’s private data at the sample level. Existing gradient inversion methods can generally be categorized into two paradigms \cite{zhang2022survey}: recursion- and iteration-based approaches.

Recursion-based attacks. Phong \etal \cite{8241854} first revealed that the input data can be successfully recovered using gradients from a shallow perceptron. Then, Fan \etal \cite{fan2020rethinking} extended to networks with convolution layers and solved the problem by converting the convolution layer into equivalent fully connected layers. Zhu \etal \cite{zhu2021rgap} further combined forward and backward propagation to reformulate the recovery task as solving a system of linear equations. Furthermore, Chen \etal \cite{chen2021understanding} then unified optimization cases under different situations and proposed a systematic optimization framework. Despite their theoretical advantage, the recursion-based methods suffer from the following limitations that impede the practicability: (1) they are typically restricted to low-resolution image reconstruction; (2) the solution framework requires global models in FL without any pooling layers or shortcut connections; (3) they are incompatible with mini-batch training; and (4) they heavily depend on valid gradients, \ie, if gradients are perturbed, most of these methods barely work.

Iteration-based attacks. In contrast to recursion-based methods that aim to derive analytical solutions, iteration-based methods formulate the recovery task as an optimization problem, minimizing the distance between the observed gradients and those computed from dummy inputs. 
Specifically, Zhu \etal \cite{zhu2019deep} first proposed to restore data samples along with the label by minimizing the MSE distance between the shared gradients and the dummy gradients generated by a pair of dummy samples. Following this line, Zhao \etal \cite{zhao2020idlg} proposed to extract the label of a single sample from the gradients before the optimization to provide more information to the reconstruction process. Subsequently, Geiping \etal \cite{geiping2020inverting} reconstructed higher-resolution images from ResNet \cite{he2016deep} by changing the distance metric and adding a fidelity regularization term. Yin \etal \cite{yin2021see} primarily focused on the recovery of larger batch sizes. By leveraging strong BN statistics and a deep pre-trained ResNet-50 as the global model (a larger model generates more gradient information), these methods can partially reconstruct certain information about input images, even under large batch sizes. Hatamizadeh \etal \cite{hatamizadeh2022gradvit} extended attacks on Vision Transformers and designed a novel token consistency regularization. 
However, these non-generative attacks generally achieve limited reconstruction performance due to the lack of explicit visual priors that capture the underlying structure and feature characteristics of images.
To this end, Jeon \etal \cite{jeon2021gradient} introduced generative priors and fine-tuned the GAN parameters to better utilize image priors and improve the quality of restored images. Based on the GAN-based research line, Li \etal \cite{li2022auditing} considered potential defense strategies in FL and proposed a new technique called gradient transformation to deal with the degraded gradients and managed to reveal private information even under strict defense strategies. 

Currently, several strong assumptions are made to provide extra information for reconstruction (\eg, known labels or BN statistics), which deviate from realistic FL settings. Nullifying these assumptions substantially degrades reconstruction performance \cite{huang2021evaluating}. In contrast, the proposed GIFD achieves superior performance without relying on such assumptions.

\subsection{GAN as prior knowledge} 
GAN model \cite{goodfellow2020generative} is a kind of deep generative model that implicitly learns the data distribution of a training set through adversarial optimization between a generator and a discriminator. A well-trained GAN can generate realistic and highly diverse images. 
Recent studies show that GAN can be leveraged to solve inverse problems \cite{xia2022gan, qiu2024closer, fang2024privacy} such as compressed sensing. 
In gradient inversion, Yin \etal \cite{yin2021see} introduced a method that utilizes a pre-trained generative model as an image prior. Jeon \etal \cite{jeon2021gradient} proposed to search the latent space and parameter space of the generative model in turn, which exploits the knowledge encoded in GAN models to reconstruct images of outstanding quality. A weakness is that GIAS requires training a specific generator for each reconstructed image, which consumes large amounts of GPU memory and inference time. Li \etal \cite{li2022auditing} also adopted the generative model, but only optimized the latent code, which achieves semantic-level reconstruction. 
Existing generative methods face two major limitations: (1) they restrict the search to the latent or parameter space of GANs, failing to fully exploit the inner prior knowledge within the GAN models; (2) their performance gains largely rely on knowledge of the private data distribution, which is unavailable in realistic FL scenarios.
\begin{figure*}
\centering
\includegraphics[width=\linewidth]{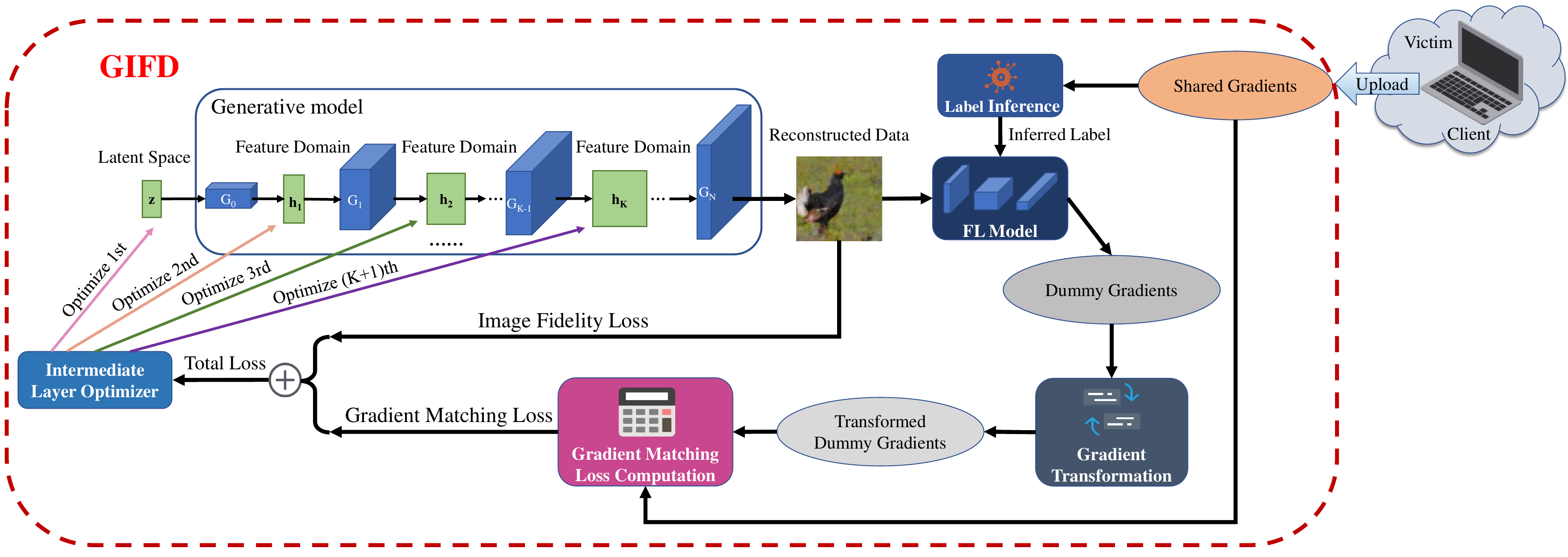}
\caption{Overview of our proposed GIFD attack. The intermediate layer optimizer minimizes the matching loss computed from the dummy gradients and the shared gradients from the victim client to update the latent vector and the intermediate features successively. 
The image fidelity regularization helps improve the quality of generated images. The reconstructed image from the layer with the corresponding least gradient matching loss is selected as the final output.} \label{pipeline_v5.pdf}
\end{figure*}
In contrast, the proposed GIFD fully exploits generative priors by searching the GAN's intermediate feature domains with sufficient knowledge. Moreover, its strong performance on OOD data reconstruction naturally relaxes the private data-distribution assumption.


%% file: method.tex
\section{METHOD}

This section first introduces the basic paradigm of gradient inversion attacks. Then, we describe how previous methods incorporate GAN to achieve better results. Finally, we illustrate our proposed GIFD, which successively searches the latent space and intermediate feature spaces of the generative model, along with various techniques for attack enhancement.

\subsection{Problem Formulation}
Given a neural network $f_\theta(\cdot)$ with parameters $\theta$ for image classification tasks, batch-averaged gradients $g$ calculated from a private batch with images $\mathbf{x^*}$ and labels $\mathbf{y^*}$, the attacker attempts to invert the gradients to recover private data with randomly initialized input tensor $\mathbf{\hat{x}} \in \mathbb{R}^{B\times H\times W \times C}$ and labels $\mathbf{\hat{y}} \in \{0,1\}^{B\times L}$ ($B,H,W,C,L$ being batch size, height, width, number of channels and class number):
\begin{equation}
\begin{aligned}
    \mathbf{\hat{x}^*}, \mathbf{\hat{y}^*} = \mathop{\arg\min}_{\mathbf{\mathbf{\hat{x}}}, \mathbf{\hat{y}}} \mathcal{D}\left(\frac{1}{B}\sum_{i=1}^{B}\nabla\ell(f_\theta(x_i),y_i), g\right),
\end{aligned}
\end{equation}
where $\mathbf{\hat{x}}=(x_1,\dots, x_B)$, $\mathbf{\hat{y}}=(y_1,\dots, y_B)$. $\mathcal{D}(\cdot,\cdot)$ is the measurement of distance, \eg, $l_2$-distance \cite{yin2021see,li2022auditing}, negative cosine similarity \cite{geiping2020inverting, jeon2021gradient}, and $\ell(\cdot,\cdot)$ is the classification loss function. In the algorithm's workflow, the attacker generates a pair of random noise $\mathbf{\hat{x}}$ and the corresponding labels $\mathbf{\hat{y}}$ as parameters, which are iteratively optimized towards the ground truth $\mathbf{x^*}$ and $\mathbf{y^*}$ through minimizing the discrepancy between dummy gradients and transmitted gradients. 


Existing studies \cite{zhao2020idlg, yin2021see, ma2022instance} have revealed that private labels can be accurately inferred by analyzing the received gradients. Therefore, the objective function with a regularization term can be simplified to the following form:

\begin{equation}
\begin{aligned}
\mathbf{\hat{x}^*}= \mathop{\arg\min}_{\mathbf{\mathbf{\hat{x}}}}\mathcal{D}\left(F(\mathbf{\hat{x}}), g\right) + R_{prior}(\mathbf{\hat{x}}),
\end{aligned}
\end{equation}
where $F(\mathbf{\hat{x}})=\frac{1}{B}\sum_{i=1}^{B}\nabla\ell(f_\theta(x_i),y_i)$, $R_{prior}(\mathbf{\hat{x}})$ is prior knowledge regularization (\eg, BN statistics \cite{yin2021see}).

Given a pre-trained generative model $G_w(\cdot)$ learning from the public dataset, an intuitive approach is to transform the problem into the following form:

\begin{equation}
\begin{aligned}
\mathbf{z^*}= \mathop{\arg\min}_{\mathbf{\mathbf{z}}}\mathcal{D}\left(F(G_w(\mathbf{z})), g\right) + R_{prior}(\mathbf{z};G_w),
\end{aligned}
\end{equation}
where $\mathbf{z} \in \mathbb{R}^{B\times k}$ is the latent code of the generative model. By narrowing the search range from $\mathbb{R}^{B\times m}$ ($m=H\times W \times C$) to a lower-dimensional subspace $\mathbb{R}^{B\times k}$ ($k << m$), one can significantly reduce the uncertainty in the optimizing process. Building on this, various GAN-based gradient inversion attacks \cite{li2022auditing,jeon2021gradient} have been proposed to enhance the quality and fidelity of generated images.


\subsection{Gradient Inversion over Feature Domains}
We first formally formulate the overall optimization objective, followed by a detailed introduction of each technique:
\begin{equation} \label{(4)}
\begin{aligned}
\mathbf{\hat{x}^*}= \mathop{\arg\min}_{\mathbf{\hat{x}}}\mathcal{D}\left(\mathcal{T}(F(\mathbf{\hat{x}})), g\right) + \mathcal{R}_{fidty}(\mathbf{\hat{x}}),
\end{aligned}
\end{equation}
where $\mathbf{\hat{x}}$ are images generated by $G_w$ or part of $G_w$, $F(\cdot)$ is the batch-averaged gradient operator, and $\mathcal{T}(\cdot)$ is the gradient transformation technique we will discuss later. 
The first term $\mathcal{D}\left(\mathcal{T}(F(\mathbf{\hat{x}})), g\right)$ denotes the gradient matching loss, while the second term, $\mathcal{R}_{fidty}(\mathbf{\hat{x}})$, serves as the fidelity regularization for realism preservation. For notational simplicity, Eq. (\ref{(4)}) is expressed in the following form:
\begin{equation}
\begin{aligned}
\mathbf{\hat{x}^*} = \mathop{\arg\min}_{\mathbf{\hat{x}}} \mathcal{L}_{grad}(\mathbf{\hat{x}}),
\end{aligned}
\end{equation}
where we denote the loss function in Eq. (\ref{(4)}) by $ \mathcal{L}_{grad}(\mathbf{\hat{x}})$. An overview of our method is shown in Figure \ref{pipeline_v5.pdf}, we next introduce each component in detail.

\noindent\textbf{Intermediate Layer Optimization.} While leveraging GAN priors has become a basic paradigm for gradient inversion, existing methods limit their search space within the latent and parameter space. Instead of intuitively optimizing the latent code \cite{li2022auditing} or fine-tuning the parameters of the GAN \cite{jeon2021gradient}, we introduce a novel feature-domain perspective and propose GIFD, the first to explicitly dissect the generator for gradient inversion and exploit the rich information encoded in its hierarchical intermediate features. As described in Algorithm \ref{GIFD}, instead of directly optimizing the dummy input $\mathbf{\hat{x}}$, we design a sequential search strategy over the latent space and the intermediate feature space of the generator to fully exploit the prior knowledge encoded in pre-trained GANs.

The first step is to optimize over the randomly initialized latent vector $\mathbf{z}$ using gradient descent, with an effective Spherical Optimizer \cite{menon2020pulse} to ensure stable convergence on the latent manifold. Once we obtain the optimal $\mathbf{z^*}$, we disassemble the generator $G_w$ into $G_0\circ G_1\circ \dots \circ G_{N-1} \circ G_N$ for intermediate feature optimization. Then, we map the optimal latent vector $\mathbf{z^*}$ into the initial intermediate latent representations $\mathbf{h_1^0}$ using $G_0$, \ie, $\mathbf{h_1^0}:=G_0(\mathbf{z^*})$. Next, the algorithm proceeds to the iterative intermediate feature search, as described in the for-loop beginning at line~\ref{line7} in Algorithm~\ref{GIFD}.

At the pass of loop $i$, we perform the following operations. First, we generate images from intermediate feature $\mathbf{h_i}$ only with the rest part of $G_w$ (\ie, $G_i\circ \dots \circ G_N$). These synthesized images are then used to compute the corresponding dummy gradients, based on which we optimize $\mathbf{h}_i$ by minimizing the loss function defined in Eq.~(\ref{(4)}). Considering the risk of producing unrealistic images during intermediate feature optimization, we constrain the search range of $\mathbf{h}_i$ within an $l_1$ ball of radius $r[i]$ centered at $\mathbf{{h_i^0}}$, \ie the term $ball_{\mathbf{h_i^0}}^{r[i]}$ in line \ref{line9} of Algorithm \ref{GIFD}. After obtaining the optimal results $\mathbf{h_i^*}$ for the current layer, we generate the initial intermediate representations for the next layer via forward propagation through $G_i$, \ie, $\mathbf{h_{i+1}^0}:=G_i(\mathbf{h_i^*})$.

As shown in lines \ref{line4}, \ref{line11}, \ref{line12}, \ref{line13}, and \ref{line18} of Algorithm \ref{GIFD}, we hope to utilize the gradient matching loss as valid information to guide us to select the output images. More specifically, among all the searched intermediate layers, we select the output corresponding to the layer that yields the lowest gradient matching loss.  
While a lower matching loss does not necessarily imply superior visual quality, our empirical results demonstrate that this adaptive selection strategy consistently outperforms outputs from a fixed layer.

\renewcommand{\algorithmicrequire}{\textbf{Input:}}  
\renewcommand{\algorithmicensure}{\textbf{Output:}} 

\begin{algorithm}[t]
  \caption{Pseudocode of our proposed GIFD} \label{GIFD}
  \begin{algorithmic}[1]
    \Require
      $G_w$: a pre-trained generative model;
      $f_\theta$: the global model in FL;
      $g$: shared gradients;
      $K$: the index of the last intermediate layer to optimize;
      $r[1\dots K]$: radius of ${l_1}$ ball in each intermediate layer;
      $B$: batch size;

    \Ensure
       Reconstructed images via GIFD attack;
       \State Initial latent code $\mathbf{z}:=(z_1, \dots , z_B)$ with random noise
            \State \parbox[t]{\dimexpr\linewidth-\algorithmicindent}{\texttt{// Latent space search\strut}}
       \State $\mathbf{z^*} \leftarrow \mathop{\arg\min}_{\mathbf{z}} \mathcal{L}_{grad}(G_w(\mathbf{z}))$ 
       \State Set $\mathbf{\hat{x}^*}:=G_w(\mathbf{z^*})$, $loss_{min} = \mathcal{D}\left(\mathcal{T}(F(G_w(\mathbf{z^*}))), g\right)$   \label{line4}
       \State Disassemble $G_w$ into $G_0\circ G_1\circ \dots \circ G_{N-1} \circ G_N$
       \State Set $\mathbf{h_1^0} := G_0(\mathbf{z^*})$ 
       \For{$i \leftarrow 1$ to $K$} \label{line7}
            
            \State \parbox[t]{\dimexpr\linewidth-\algorithmicindent}{\texttt{//Intermediate layers search with $l_1$-ball constraint\strut}}

        \State $\mathbf{h_i^*} \leftarrow argmin_{\mathbf{h_i}\in  ball_{\mathbf{h_i^0}}^{r[i]}}\mathcal{L}_{grad}(G_i\circ \dots \circ G_N(\mathbf{h_i})) $  \label{line9}
        \State $loss_i = \mathcal{D}\left(\mathcal{T}(F(G_i\circ \dots \circ G_N(\mathbf{h_i^*}))), g\right)$
      \If {$loss_i<loss_{min}$}  \label{line11} \State $\mathbf{\hat{x}^*} := G_i\circ \dots \circ G_N(\mathbf{h_i^*})$  \label{line12}
      \State $loss_{min}=loss_i$  \label{line13}
     \EndIf
     \State \parbox[t]{\dimexpr\linewidth-\algorithmicindent}{\texttt{// Generate features of the next intermediate layer as the initial vector to optimize\strut}}

       \State $\mathbf{h_{i+1}^0} := G_i(\mathbf{h_i^*}) $
     \EndFor
  \State Return results: $\mathbf{\hat{x}^*}$ \label{line18}
  \end{algorithmic}
\end{algorithm}

With all the efforts above, we encourage the optimizer to explore the intermediate space with rich information to generate more diverse and high-fidelity images, while limiting the solution space within an $l_1$ ball around the manifold induced by the previous layer, to avoid overfitting and guarantee the realism of the generated images. Notably, our approach is easy to implement, as it is agnostic to specific GAN architectures and only requires access to a pre-trained generative model.


\noindent\textbf{Image Fidelity Regularization.} Given that the gradients are only a non-linear transformation of the original data, it is highly challenging to restore data only from shared gradients. This motivates the incorporation of strong image priors to approximate the underlying data distribution and enhance reconstruction fidelity:
\begin{equation}
\begin{aligned}
\mathcal{R}_{fidty}(\mathbf{\hat{x}}) = \alpha_{\ell_2}\mathcal{R}_{\ell_2}(\mathbf{\hat{x}})+\alpha_{TV}\mathcal{R}_{TV}(\mathbf{\hat{x}}),
\end{aligned}
\end{equation}
where the first term is the $\ell_2$ norm of the images with scaling factor $\alpha_{\ell_2}$, which encourages the algorithm to solve for a solution that is preferably sparse. To further promote spatial smoothness, we incorporate a total variation (TV) regularization term $\mathcal{R}_{TV}(\mathbf{\hat{x}})$ \cite{geiping2020inverting} scaled by $\alpha_{TV}$, which penalizes abrupt intensity changes between neighboring pixels, in line with the natural statistics of images.

\noindent\textbf{Gradient Transformation.} To mitigate the effects of several defense strategies based on gradient perturbation, we apply the adaptive strategy proposed in \cite{li2022auditing} to all methods for fair comparison, \ie, $\mathcal{T}(\cdot)$ in Eq. (\ref{(4)}). In practice, an attacker may infer the deployed defense and estimate its hyperparameters by dynamically observing the received gradients. \eg, the proportion of zero-valued gradients can indicate the use of sparsification and provide an estimate of its sparsity level. Specifically, the algorithm is able to infer three defense strategies: (1) \textit{Gradient clipping}; (2) \textit{Gradient sparsification}; and (3) \textit{Soteria} \cite{sun2021soteria}, as detailed in Appendix A. 

\noindent\textbf{Label Extraction.} Consider a network parameterized by $\mathbf{\rm W}$ for a classification task over $n$-classes using cross-entropy loss, when the training data is a single image, the ground truth label $c$ can be accurately inferred \cite{zhao2020idlg} through:
\begin{equation}
\begin{aligned}
 c = i, \ \ \ {\rm s.t.} \ \nabla \mathbf{{\rm W}_{FC}^i{^{\top}}} \cdot \nabla \mathbf{{\rm W}_{FC}^j} \leq 0, \ \forall\ j \neq i,
 \label{(7)}
\end{aligned}
\end{equation}
where we denote the gradient vector \wrt the weights (denoted as $\mathbf{{\rm W}_{FC}^i}$) connected to the $i_{th}$ logit in the classification layer (\ie, the output layer) by $\mathbf{\nabla {\rm W}_{FC}^i}$. Eq. (\ref{(7)}) identifies the ground-truth label via the index of the negative gradients. Yin \etal \cite{yin2021see} extended to support batch-level label extraction but assumed non-repeating labels in the batch. Ma \etal \cite{ma2022instance} further formulated it as a problem of solving linear equations, which reaches 99\% accuracy without relying on impractical assumptions. The inferred labels are used to compute dummy gradients and as class conditions for GANs, which greatly enhance our attack.

\begin{figure}[t]
\centering
\includegraphics[width=\linewidth]{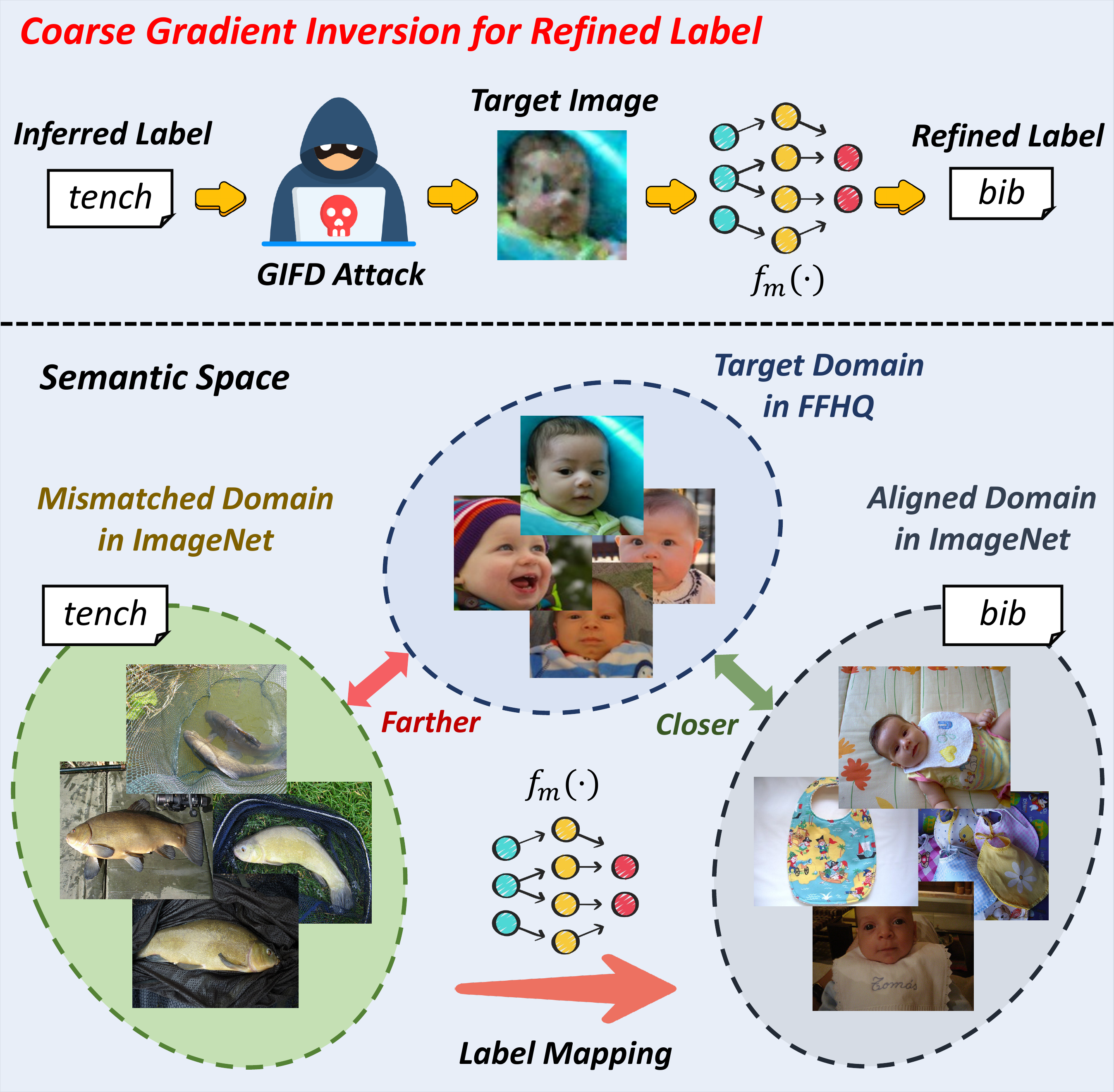}
\caption{An illustration of our proposed label mapping mechanism. We first conduct only a small number of iterations based on the initially inferred label for a coarse reconstruction $\mathbf{\hat{x}}$. It is then fed into $f_m(\cdot)$ for a refined label, which serves as the conditional input for GAN models in subsequent fine-grained attacks. This technique corrects the label inconsistency and offers more aligned guidance in the semantic space.} 
\label{fig:label_mapping}
\end{figure}

\subsection{Label Mapping for Label Inconsistency}
In class-conditioned image generation, label information is injected into conditional GANs to specify the target class and control the synthesis process. However, in more practical OOD scenarios, the semantic meanings of label indices in the GAN's training dataset and the FL model's training dataset may differ significantly.
The attacker can only obtain the label index of the private image through label extraction, which may not align with the GAN’s label semantics, leading to misleading conditional guidance during inversion.
For instance, when attacking a human face classifier trained on FFHQ using a conditional GAN (e.g., BigGAN) pretrained on ImageNet, the attacker may infer that the label index of the private image is $0$. However, index $0$ corresponds to ``tench" in ImageNet, while representing people in the age group $[0, 9]$ in FFHQ. As a result, the incorrect label information can degrade the reconstruction quality and hinder effective inversion.

To address this label mismatch, we introduce a label-mapping mechanism that enhances the reliability of label guidance under such OOD conditions.
The key intuition is that a well-trained classifier encapsulates comprehensive knowledge about the semantics of different categories and is expected to determine the most plausible label for a given target image.
Specifically, we first train a classifier $f_{m}(\cdot)$ on the GAN's training dataset, which is assumed to be accessible to the adversary.
Next, we disassemble the inversion into two sequential stages. In the first stage, a coarse gradient inversion is conducted using the initially inferred (and potentially inadequate) label, requiring only a small number of iterations to obtain a rough reconstruction $\mathbf{\hat{x}}'$.
Although $\mathbf{\hat{x}}'$ is not visually perfect, it often preserves sufficient high-level semantic information for subsequent category prediction.
As illustrated in Figure \ref{fig:label_mapping}, $\mathbf{\hat{x}}'$ is then fed into the trained classifier $f_m(\cdot)$ to derive a refined label prediction $y_{\text{rec}}$, which is then used as the new conditioning input for the generator in a fine-grained inversion process.
Essentially, the classifier $f_m(\cdot)$ serves as a semantic remapper, which corrects the potentially mismatched label to one that better aligns with the GAN’s semantic space, hence alleviating the adverse effects of label inconsistency during inversion.
Consequently, when the mapped label provides a conditional prior with greater semantic overlap with the target image, the subsequent fine-grained inversion can benefit more from the corrected semantic guidance, leading to greater reconstruction gains.

%% file: Experiments.tex
\section{Experiments}
\label{noise}

To validate the effectiveness of GIFD in improving attack performance, we conduct experiments on two widely used GANs in a range of scenarios. We evaluate our method for the classification task on the validation set of the ImageNet ILSVRC 2012 dataset \cite{deng2009imagenet} and 10-class (using age as label) FFHQ \cite{karras2019style} at $64\times64$ pixels. ImageNet ILSVRC 2012 is a large-scale, generic dataset with diverse object categories and substantial visual variation, which is well-suited for evaluating the generalization and robustness of GLA. FFHQ provides high-quality and diverse human face images, serving as a complementary domain for evaluating whether the proposed method can reconstruct realistic, structurally complex data beyond general objects. The combination of the two datasets allows us to evaluate attacks across different visual domains and assess their generalization capabilities comprehensively. For the generative model, we use a pre-trained BigGAN \cite{brock2018large} for ImageNet and a pre-trained StyleGAN2 \cite{karras2019style} for FFHQ. We use a randomly initialized ResNet-18 as the default FL model and choose negative cosine similarity as the distance metric $\mathcal{D}(\cdot)$. Following existing works \cite{geiping2020inverting, jeon2021gradient}, we use the default $B=1$ at one local step. Then we conduct experiments with larger $B$ and compare the performance of different methods. Our code is available at \textcolor{magenta}{\url{https://github.com/ffhibnese/GIFD}}.

\noindent\textbf{Implementation details}. According to its specific structure, we split BigGAN into $G_0$ to $G_{12}$ with 12 intermediate feature domains, and StyleGAN2 into $G_0$ to $G_7$ with 7 intermediate feature domains. We ensure that the intermediate features lie in the $l_1$ ball through Project Gradient Descent (PGD) \cite{nesterov2003introductory}. Motivated by the fact that a stepwise optimization over the noises in StyleGAN2 yields better reconstructions \cite{daras2021intermediate} for compressed sensing, we gradually allow to optimize more noise as we move to deeper intermediate layers and make them lie inside the $l_1$ ball as well. 
For each intermediate feature domain, we use the Adam optimizer with $0.1$ as the initial learning rate and give $1000$ iterations. In the coarse inversion stage of label mapping for refined labels, we only give $100$ iterations for each feature layer. We adopt the warm-up strategy, where the learning rate linearly warms up from $0$ to $0.1$ during the first $1/20$ of the optimization and gradually decays to $0$ in the last $3/4$ stage using cosine decay.
For the image fidelity regularization, we use $\alpha_ {TV} = 10^{-4}$ and $\alpha_{l_2} = 10^{-6}$. The experiments on the effects of K and on defense strategies are each conducted on 30 randomly selected images, and the numerical results for batch size are the averages of 10 batches. We run all experiments on NVIDIA RTX 2080 Ti GPUs and A100 GPUs. More details are provided in Appendix B.

\noindent\textbf{Evaluation Metrics}. We compute the following metrics to measure the discrepancy between reconstructed and ground-truth images: (1) PSNR (Peak Signal-to-Noise Ratio), which quantifies the pixel-level fidelity of reconstructed images by measuring the ratio between the signal strength and reconstruction error; (2) LPIPS \cite{zhang2018unreasonable} (Learned Perceptual Image Patch Similarity), which measures perceptual similarity based on deep feature representations from pretrained networks; (3) SSIM (Structural Similarity Index Measure), which evaluates image similarity in terms of luminance, contrast, and structural information; and (4) MSE (Mean Squared Error), which measures the average squared pixel-wise reconstruction error. Higher values of PSNR and SSIM indicate better reconstruction quality, and the opposite for LPIPS and MSE.

\begin{figure}[t]
	\centering
	\begin{subfigure}{0.49\linewidth}
		\centering
		\includegraphics[width=0.95\linewidth]{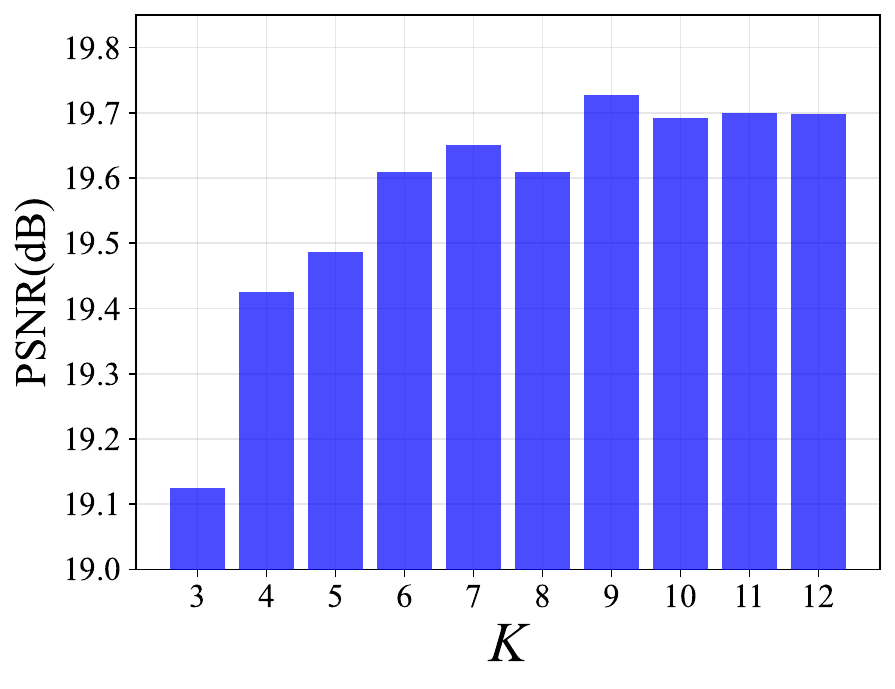}
		\caption{ImageNet (BigGAN)}
		\label{select layer}
	\end{subfigure}
	\begin{subfigure}{0.49\linewidth}
		\centering
		\includegraphics[width=0.95\linewidth]{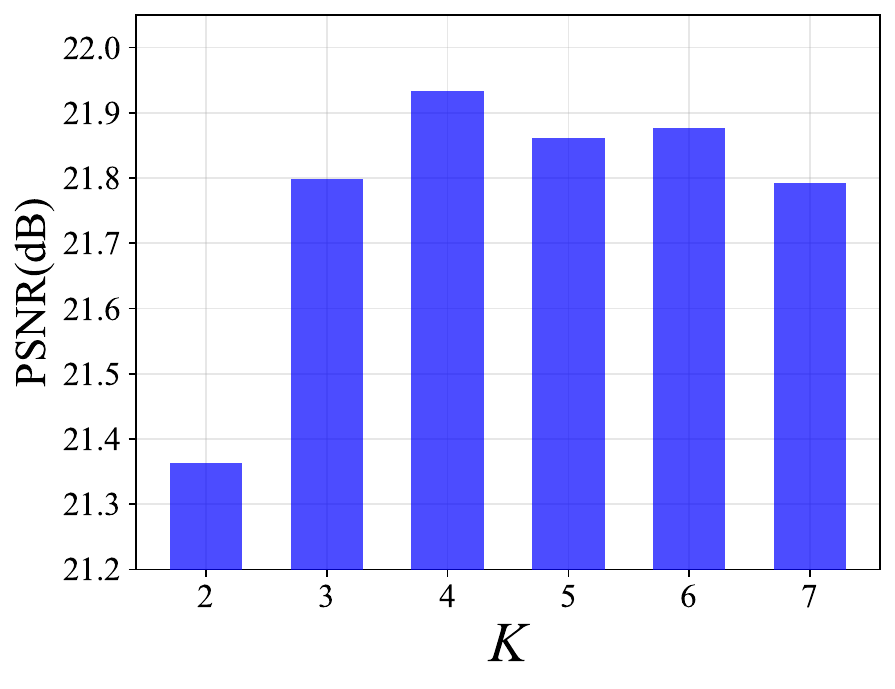}
		\caption{FFHQ (StyleGAN2)}
		\label{select layer ffhq}
	\end{subfigure}
	\centering
	\caption{Quantitative results on ImageNet (BigGAN) and FFHQ (StyleGAN2) under different values of $K$ in Algorithm \ref{GIFD}. Notably, the figures exclude the results where the corresponding values are below the starting point of the y-axis.}
	\label{select_layer_psnr}
\end{figure}

\begin{figure}[t]
\centering

\begin{subfigure}{\linewidth}
\centering

\begin{minipage}[t]{0.158\linewidth}\centering
\includegraphics[width=1.4cm]{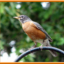}
\end{minipage}\hfill%
\begin{minipage}[t]{0.158\linewidth}\centering
\includegraphics[width=1.4cm]{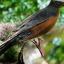}
\end{minipage}\hfill%
\begin{minipage}[t]{0.158\linewidth}\centering
\includegraphics[width=1.4cm]{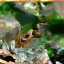}
\end{minipage}\hfill%
\begin{minipage}[t]{0.158\linewidth}\centering
\includegraphics[width=1.4cm]{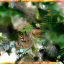}
\end{minipage}\hfill%
\begin{minipage}[t]{0.158\linewidth}\centering
\includegraphics[width=1.4cm]{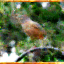}
\end{minipage}\hfill%
\begin{minipage}[t]{0.158\linewidth}\centering
\includegraphics[width=1.4cm]{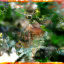}
\end{minipage}


\begin{minipage}[t]{0.158\linewidth}\centering
\includegraphics[width=1.4cm]{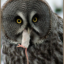}
\end{minipage}\hfill%
\begin{minipage}[t]{0.158\linewidth}\centering
\includegraphics[width=1.4cm]{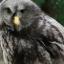}
\end{minipage}\hfill%
\begin{minipage}[t]{0.158\linewidth}\centering
\includegraphics[width=1.4cm]{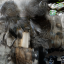}
\end{minipage}\hfill%
\begin{minipage}[t]{0.158\linewidth}\centering
\includegraphics[width=1.4cm]{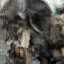}
\end{minipage}\hfill%
\begin{minipage}[t]{0.158\linewidth}\centering
\includegraphics[width=1.4cm]{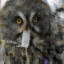}
\end{minipage}\hfill%
\begin{minipage}[t]{0.158\linewidth}\centering
\includegraphics[width=1.4cm]{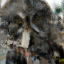}
\end{minipage}

\begin{minipage}[t]{0.158\linewidth}\centering\scriptsize Original\end{minipage}\hfill%
\begin{minipage}[t]{0.158\linewidth}\centering\scriptsize $K{=}0$\end{minipage}\hfill%
\begin{minipage}[t]{0.158\linewidth}\centering\scriptsize $K{=}3$\end{minipage}\hfill%
\begin{minipage}[t]{0.158\linewidth}\centering\scriptsize $K{=}6$\end{minipage}\hfill%
\begin{minipage}[t]{0.158\linewidth}\centering\scriptsize $K{=}9$\end{minipage}\hfill%
\begin{minipage}[t]{0.158\linewidth}\centering\scriptsize $K{=}12$\end{minipage}

\caption{ImageNet (BigGAN)}
\label{k_vis_imagenet}
\end{subfigure}%
\hfill%

\vspace{1.5mm}

\begin{subfigure}{\linewidth}
\centering

\begin{minipage}[t]{0.158\linewidth}\centering
\includegraphics[width=1.4cm]{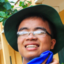}
\end{minipage}\hfill%
\begin{minipage}[t]{0.158\linewidth}\centering
\includegraphics[width=1.4cm]{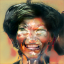}
\end{minipage}\hfill%
\begin{minipage}[t]{0.158\linewidth}\centering
\includegraphics[width=1.4cm]{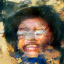}
\end{minipage}\hfill%
\begin{minipage}[t]{0.158\linewidth}\centering
\includegraphics[width=1.4cm]{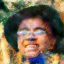}
\end{minipage}\hfill%
\begin{minipage}[t]{0.158\linewidth}\centering
\includegraphics[width=1.4cm]{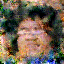}
\end{minipage}\hfill%
\begin{minipage}[t]{0.158\linewidth}\centering
\includegraphics[width=1.4cm]{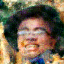}
\end{minipage}


\begin{minipage}[t]{0.158\linewidth}\centering
\includegraphics[width=1.4cm]{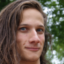}
\end{minipage}\hfill%
\begin{minipage}[t]{0.158\linewidth}\centering
\includegraphics[width=1.4cm]{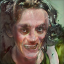}
\end{minipage}\hfill%
\begin{minipage}[t]{0.158\linewidth}\centering
\includegraphics[width=1.4cm]{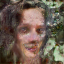}
\end{minipage}\hfill%
\begin{minipage}[t]{0.158\linewidth}\centering
\includegraphics[width=1.4cm]{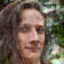}
\end{minipage}\hfill%
\begin{minipage}[t]{0.158\linewidth}\centering
\includegraphics[width=1.4cm]{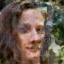}
\end{minipage}\hfill%
\begin{minipage}[t]{0.158\linewidth}\centering
\includegraphics[width=1.4cm]{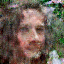}
\end{minipage}

\begin{minipage}[t]{0.158\linewidth}\centering\scriptsize Original\end{minipage}\hfill%
\begin{minipage}[t]{0.158\linewidth}\centering\scriptsize $K{=}0$\end{minipage}\hfill%
\begin{minipage}[t]{0.158\linewidth}\centering\scriptsize $K{=}2$\end{minipage}\hfill%
\begin{minipage}[t]{0.158\linewidth}\centering\scriptsize $K{=}4$\end{minipage}\hfill%
\begin{minipage}[t]{0.158\linewidth}\centering\scriptsize $K{=}6$\end{minipage}\hfill%
\begin{minipage}[t]{0.158\linewidth}\centering\scriptsize $K{=}8$\end{minipage}

\caption{FFHQ (StyleGAN2)}
\label{k_vis_ffhq}
\end{subfigure}

\vspace{-1.0mm}

\caption{Qualitative results of the reconstructed images under different values of $K$ in Algorithm \ref{GIFD}. Each row corresponds to one sample. The best visual quality is attained at $K=9$ for ImageNet (BigGAN) and $K=4$ for FFHQ (StyleGAN2), which is consistent with the quantitative results in Figure~\ref{select_layer_psnr}.}
\label{fig:k_selection_visualization}
\end{figure}

\subsection{Decide Which Layer to End}
To further improve the quality of output images, we need to carefully handle the parameter $K$ in Algorithm \ref{GIFD}. Experimentally, we find that there is a trade-off between under-fitting and over-fitting regarding the choice of $K$. When $K$ is small, we only search the first few intermediate features of the generative model and do not fully utilize the rich information encoded in the intermediate space. As a result, the quality of the generated images does not meet our expectations. On the contrary, when $K$ is large, we excessively search the deeper layers and generate images with lower values of the cost function but a larger discrepancy from the original images. Therefore, we randomly select images (disjoint from our main experimental data) from the validation set of ImageNet and FFHQ to study the impact of $K$ and try to select the best final layer. As shown in Figure \ref{select_layer_psnr}, when $K=9$ and $K=4$ are used for BigGAN and StyleGAN2, respectively, we obtain results with the largest PSNR. The reconstructed images in Figure \ref{fig:k_selection_visualization} also provide intuitive evidence for our choices.

\begin{table*}[htbp]
  \centering
  \caption{Comparison of GIFD with state-of-the-art methods on every 1000th image of the ImageNet and FFHQ validation sets. We calculate the average value of metrics on reconstructed images.}
    \resizebox{17cm}{!}{\begin{tabular}{ccccccccccc} \toprule
    \multicolumn{1}{c}{\multirow{2}[0]{*}{Metric}} & \multicolumn{5}{c}{ImageNet}          & \multicolumn{5}{c}{FFHQ} \\ \cmidrule(lr){2-6} \cmidrule(lr){7-11} 
    \multicolumn{1}{c}{} & IG \cite{geiping2020inverting}    & GI \cite{yin2021see}    & GGL \cite{li2022auditing}   & GIAS \cite{jeon2021gradient}  & \textbf{GIFD}  & \multicolumn{1}{l}{IG \cite{geiping2020inverting}} & \multicolumn{1}{l}{GI \cite{yin2021see}} & \multicolumn{1}{l}{GGL \cite{li2022auditing}} & \multicolumn{1}{l}{GIAS \cite{jeon2021gradient}} & \multicolumn{1}{l}{\textbf{GIFD}} \\ \midrule
    PSNR$\uparrow$  & 17.0756  & 16.5109  & 13.3885  & 17.4923  & \textbf{20.0534 } & 15.3523  & 14.9485  & 15.1335  & 20.1799  & \textbf{21.3368} \\
    LPIPS$\downarrow$ & 0.3078  & 0.3297  & 0.3678  & 0.2536  & \textbf{0.1559 } & 0.4172  & 0.4503  & 0.2009  & 0.1266  & \textbf{0.1023 } \\
    SSIM$\uparrow$  & 0.2908  & 0.2673  & 0.1251  & 0.3381  & \textbf{0.4713 } & 0.2272  & 0.2044  & 0.2453  & 0.5379  & \textbf{0.5768} \\
    MSE$\downarrow$  & 0.0223  & 0.0258  & 0.0553  & 0.0236  & \textbf{0.0141 } & 0.0311  & 0.0343  & 0.0339  & 0.0121  & \textbf{0.0098} \\  \bottomrule
    \end{tabular}}
  \label{main_tab}%
\end{table*}

\begin{figure*}[htbp]
    \centering
    \begin{subfigure}{0.48\linewidth}
        \begin{minipage}[t]{0.01\linewidth}
        \rotatebox{90}{\scriptsize{\textbf{}}}     
        \end{minipage}
		\begin{minipage}[t]{0.165\linewidth}
        \centering
        \includegraphics[width=1.4cm]{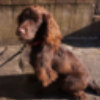}
        \end{minipage}%
        \begin{minipage}[t]{0.165\linewidth}
        \centering
        \includegraphics[width=1.4cm]{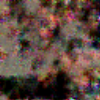}
        \end{minipage}%
        \begin{minipage}[t]{0.165\linewidth}
        \centering
        \includegraphics[width=1.4cm]{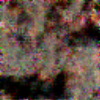}
        \end{minipage}%
        \begin{minipage}[t]{0.165\linewidth}
        \centering
        \includegraphics[width=1.4cm]{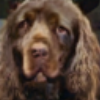}
        \end{minipage}%
        \begin{minipage}[t]{0.165\linewidth}
        \centering
        \includegraphics[width=1.4cm]{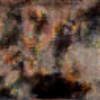}
        \end{minipage}%
        \begin{minipage}[t]{0.165\linewidth}
        \centering
        \includegraphics[width=1.4cm]{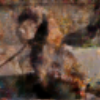}
        \end{minipage}%
        \qquad 
        \begin{minipage}[t]{0.01\linewidth}
        \rotatebox{90}{\scriptsize{\textbf{}}}     
        \end{minipage}
        \begin{minipage}[t]{0.165\linewidth}
        \centering
        \includegraphics[width=1.4cm]{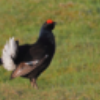}
        \end{minipage}%
        \begin{minipage}[t]{0.165\linewidth}
        \centering
        \includegraphics[width=1.4cm]{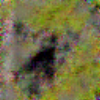}
        \end{minipage}%
        \begin{minipage}[t]{0.165\linewidth}
        \centering
        \includegraphics[width=1.4cm]{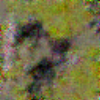}
        \end{minipage}%
        \begin{minipage}[t]{0.165\linewidth}
        \centering
        \includegraphics[width=1.4cm]{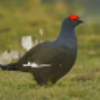}
        \end{minipage}%
        \begin{minipage}[t]{0.165\linewidth}
        \centering
        \includegraphics[width=1.4cm]{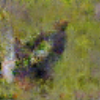}
        \end{minipage}%
        \begin{minipage}[t]{0.165\linewidth}
        \centering
        \includegraphics[width=1.4cm]{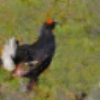}
        \end{minipage}
        \qquad 
        \begin{minipage}[t]{0.01\linewidth}
        \rotatebox{90}{\scriptsize{\textbf{}}}     
        \end{minipage}
        \begin{minipage}[t]{0.165\linewidth}
        \centering
        \includegraphics[width=1.4cm]{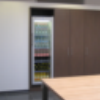}
        \end{minipage}%
        \begin{minipage}[t]{0.165\linewidth}
        \centering
        \includegraphics[width=1.4cm]{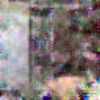}
        \end{minipage}%
        \begin{minipage}[t]{0.165\linewidth}
        \centering
        \includegraphics[width=1.4cm]{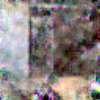}
        \end{minipage}%
        \begin{minipage}[t]{0.165\linewidth}
        \centering
        \includegraphics[width=1.4cm]{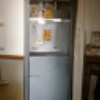}
        \end{minipage}%
        \begin{minipage}[t]{0.165\linewidth}
        \centering
        \includegraphics[width=1.4cm]{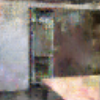}
        \end{minipage}%
        \begin{minipage}[t]{0.165\linewidth}
        \centering
        \includegraphics[width=1.4cm]{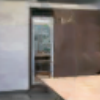}
        \end{minipage}
        \qquad 
        \begin{minipage}[t]{0.01\linewidth}
        \rotatebox{90}{\scriptsize{\textbf{}}}     
        \end{minipage}
        \begin{minipage}[t]{0.165\linewidth}
        \centering
        \includegraphics[width=1.4cm]{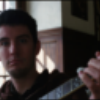}
        \end{minipage}%
        \begin{minipage}[t]{0.165\linewidth}
        \centering
        \includegraphics[width=1.4cm]{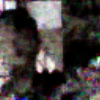}
        \end{minipage}%
        \begin{minipage}[t]{0.165\linewidth}
        \centering
        \includegraphics[width=1.4cm]{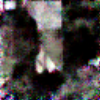}
        \end{minipage}%
        \begin{minipage}[t]{0.165\linewidth}
        \centering
        \includegraphics[width=1.4cm]{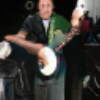}
        \end{minipage}%
        \begin{minipage}[t]{0.165\linewidth}
        \centering
        \includegraphics[width=1.4cm]{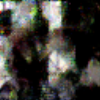}
        \end{minipage}%
        \begin{minipage}[t]{0.165\linewidth}
        \centering
        \includegraphics[width=1.4cm]{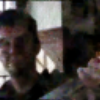}
        \end{minipage}
        \qquad 
        \begin{minipage}[t]{0.01\linewidth}
        \rotatebox{90}{\scriptsize{\textbf{}}}     
        \end{minipage}
        \begin{minipage}[t]{0.165\linewidth}
        \centering
        \includegraphics[width=1.4cm]{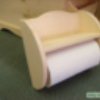}
        \end{minipage}%
        \begin{minipage}[t]{0.165\linewidth}
        \centering
        \includegraphics[width=1.4cm]{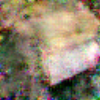}
        \end{minipage}%
        \begin{minipage}[t]{0.165\linewidth}
        \centering
        \includegraphics[width=1.4cm]{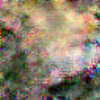}
        \end{minipage}%
        \begin{minipage}[t]{0.165\linewidth}
        \centering
        \includegraphics[width=1.4cm]{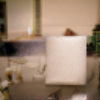}
        \end{minipage}%
        \begin{minipage}[t]{0.165\linewidth}
        \centering
        \includegraphics[width=1.4cm]{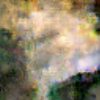}
        \end{minipage}%
        \begin{minipage}[t]{0.165\linewidth}
        \centering
        \includegraphics[width=1.4cm]{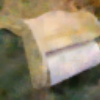}
        \end{minipage}%
        \qquad
        \begin{minipage}[t]{0.01\linewidth}
        \rotatebox{90}{\scriptsize{\textbf{}}}     
        \end{minipage}
        \begin{minipage}[t]{0.165\linewidth}
        \centering
        \includegraphics[width=1.4cm]{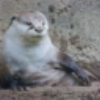}
        \centering
        \caption*{\textbf{\footnotesize{Original}}}
        \end{minipage}%
        \begin{minipage}[t]{0.165\linewidth}
        \centering
        \includegraphics[width=1.4cm]{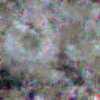}
        \centering
        \caption*{\textbf{\footnotesize{IG \cite{geiping2020inverting}}}}
        \end{minipage}%
        \begin{minipage}[t]{0.165\linewidth}
        \centering
        \includegraphics[width=1.4cm]{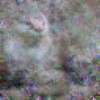}
        \centering
        \caption*{\textbf{\footnotesize{GI \cite{yin2021see}}}}
        \end{minipage}%
        \begin{minipage}[t]{0.165\linewidth}
        \centering
        \includegraphics[width=1.4cm]{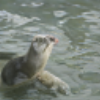}
        \centering
        \caption*{\textbf{\footnotesize{GGL \cite{li2022auditing}}}}
        \end{minipage}%
        \begin{minipage}[t]{0.165\linewidth}
        \centering
        \includegraphics[width=1.4cm]{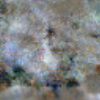}
        \centering
        \caption*{\textbf{\footnotesize{GIAS \cite{jeon2021gradient}}}}
        \end{minipage}%
        \begin{minipage}[t]{0.165\linewidth}
        \centering
        \includegraphics[width=1.4cm]{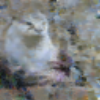}
        \centering
        \caption*{\textbf{\footnotesize{GIFD}}}
        \end{minipage}
	\caption{ImageNet (BigGAN)}
        \end{subfigure}
	\centering
	\begin{subfigure}{0.48\linewidth}
		\begin{minipage}[t]{0.165\linewidth}
        \centering
        \includegraphics[width=1.4cm]{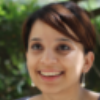}
        \end{minipage}%
        \begin{minipage}[t]{0.165\linewidth}
        \centering
        \includegraphics[width=1.4cm]{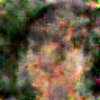}
        \end{minipage}%
        \begin{minipage}[t]{0.165\linewidth}
        \centering
        \includegraphics[width=1.4cm]{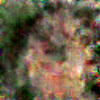}
        \end{minipage}%
        \begin{minipage}[t]{0.165\linewidth}
        \centering
        \includegraphics[width=1.4cm]{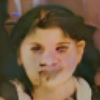}
        \end{minipage}%
        \begin{minipage}[t]{0.165\linewidth}
        \centering
        \includegraphics[width=1.4cm]{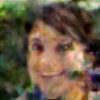}
        \end{minipage}%
        \begin{minipage}[t]{0.165\linewidth}
        \centering
        \includegraphics[width=1.4cm]{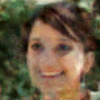}
        \end{minipage}%
        \qquad 
        \begin{minipage}[t]{0.165\linewidth}
        \centering
        \includegraphics[width=1.4cm]{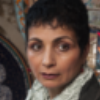}
        \end{minipage}%
        \begin{minipage}[t]{0.165\linewidth}
        \centering
        \includegraphics[width=1.4cm]{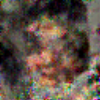}
        \end{minipage}%
        \begin{minipage}[t]{0.165\linewidth}
        \centering
        \includegraphics[width=1.4cm]{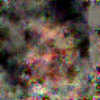}
        \end{minipage}%
        \begin{minipage}[t]{0.165\linewidth}
        \centering
        \includegraphics[width=1.4cm]{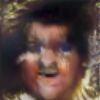}
        \end{minipage}%
        \begin{minipage}[t]{0.165\linewidth}
        \centering
        \includegraphics[width=1.4cm]{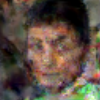}
        \end{minipage}%
        \begin{minipage}[t]{0.165\linewidth}
        \centering
        \includegraphics[width=1.4cm]{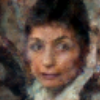}
        \end{minipage}
        \qquad 
        \begin{minipage}[t]{0.165\linewidth}
        \centering
        \includegraphics[width=1.4cm]{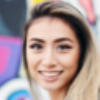}
        \end{minipage}%
        \begin{minipage}[t]{0.165\linewidth}
        \centering
        \includegraphics[width=1.4cm]{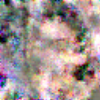}
        \end{minipage}%
        \begin{minipage}[t]{0.165\linewidth}
        \centering
        \includegraphics[width=1.4cm]{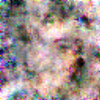}
        \end{minipage}%
        \begin{minipage}[t]{0.165\linewidth}
        \centering
        \includegraphics[width=1.4cm]{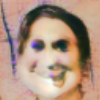}
        \end{minipage}%
        \begin{minipage}[t]{0.165\linewidth}
        \centering
        \includegraphics[width=1.4cm]{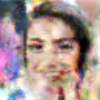}
        \end{minipage}%
        \begin{minipage}[t]{0.165\linewidth}
        \centering
        \includegraphics[width=1.4cm]{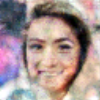}
        \end{minipage}
        \qquad 
        \begin{minipage}[t]{0.165\linewidth}
        \centering
        \includegraphics[width=1.4cm]{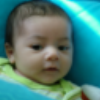}
        \end{minipage}%
        \begin{minipage}[t]{0.165\linewidth}
        \centering
        \includegraphics[width=1.4cm]{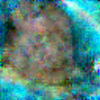}
        \end{minipage}%
        \begin{minipage}[t]{0.165\linewidth}
        \centering
        \includegraphics[width=1.4cm]{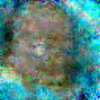}
        \end{minipage}%
        \begin{minipage}[t]{0.165\linewidth}
        \centering
        \includegraphics[width=1.4cm]{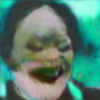}
        \end{minipage}%
        \begin{minipage}[t]{0.165\linewidth}
        \centering
        \includegraphics[width=1.4cm]{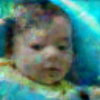}
        \end{minipage}%
        \begin{minipage}[t]{0.165\linewidth}
        \centering
        \includegraphics[width=1.4cm]{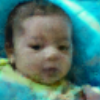}
        \end{minipage}
        \qquad 
        \begin{minipage}[t]{0.165\linewidth}
        \centering
        \includegraphics[width=1.4cm]{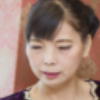}
        \end{minipage}%
        \begin{minipage}[t]{0.165\linewidth}
        \centering
        \includegraphics[width=1.4cm]{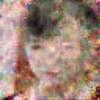}
        \end{minipage}%
        \begin{minipage}[t]{0.165\linewidth}
        \centering
        \includegraphics[width=1.4cm]{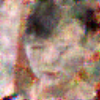}
        \end{minipage}%
        \begin{minipage}[t]{0.165\linewidth}
        \centering
        \includegraphics[width=1.4cm]{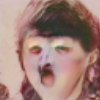}
        \end{minipage}%
        \begin{minipage}[t]{0.165\linewidth}
        \centering
        \includegraphics[width=1.4cm]{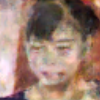}
        \end{minipage}%
        \begin{minipage}[t]{0.165\linewidth}
        \centering
        \includegraphics[width=1.4cm]{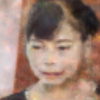}
        \end{minipage}
        \qquad
        \begin{minipage}[t]{0.165\linewidth}
        \centering
        \includegraphics[width=1.4cm]{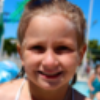}
        \centering
        \caption*{\textbf{\footnotesize{Original}}}
        \end{minipage}%
        \begin{minipage}[t]{0.165\linewidth}
        \centering
        \includegraphics[width=1.4cm]{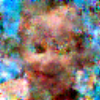}
        \centering
        \caption*{\textbf{\footnotesize{IG \cite{geiping2020inverting}}}}
        \end{minipage}%
        \begin{minipage}[t]{0.165\linewidth}
        \centering
        \includegraphics[width=1.4cm]{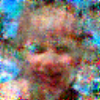}
        \centering
        \caption*{\textbf{\footnotesize{GI \cite{yin2021see}}}}
        \end{minipage}%
        \begin{minipage}[t]{0.165\linewidth}
        \centering
        \includegraphics[width=1.4cm]{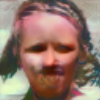}
        \centering
        \caption*{\textbf{\footnotesize{GGL \cite{li2022auditing}}}}
        \end{minipage}%
        \begin{minipage}[t]{0.165\linewidth}
        \centering
        \includegraphics[width=1.4cm]{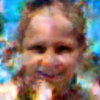}
                \centering
        \caption*{\textbf{\footnotesize{GIAS \cite{jeon2021gradient}}}}
        \end{minipage}%
        \begin{minipage}[t]{0.165\linewidth}
        \includegraphics[width=1.4cm]{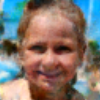}
        \centering
        \caption*{\textbf{\footnotesize{GIFD}}}
        \end{minipage}
	
            \caption{FFHQ (StyleGAN2)}
        \end{subfigure}
	\centering

	\caption{Qualitative results of different methods on ImageNet and FFHQ.}
	\label{main_figure}
\end{figure*}

\subsection{Comparison with the State-of-the-art Attacks}

Next, we compare our proposed GIFD with existing methods and provide qualitative and quantitative results. We consider the following four state-of-the-art baselines: (1) \textit{Inverting Gradients (IG)} by Geiping \etal  \cite{geiping2020inverting}; (2) \textit{GradInversion (GI)} by Yin \etal  \cite{yin2021see}; (3) \textit{Gradient Inversion in Alternative Spaces (GIAS)} by Jeon \etal  \cite{jeon2021gradient}; and (4) \textit{Generative Gradient Leakage (GGL)} by Li \etal  \cite{li2022auditing}.

In real application scenarios, a vast majority of FL systems do not transmit the BN statistics computed from private data \cite{huang2021evaluating}. Based on this fact, none of the experiments use the strong BN prior proposed by  \cite{yin2021see}. Since the randomly initialized values of vectors will greatly affect the reconstruction results, we conduct 4 trials for every attack and select the result with the least gradient matching loss.

\noindent\textbf{Experiment Results.} By observing the results in Table \ref{main_tab}, we demonstrate that the proposed GIFD consistently achieves a remarkable improvement compared to the competing baselines. Especially in the ImageNet dataset with BigGAN, our method shows nearly 2.5dB and 0.1 improvements in average PSNR and LPIPS values, respectively. As the visual comparison shown in Figure \ref{main_figure}, under a more practical setting, most existing methods struggle to recover meaningful and high-quality images even with $B=1$. In contrast, our method successfully reveals substantial information about the private data and achieves fine-grained reconstruction on both datasets.

The GAN-based methods (i.e., GGL, GIAS, GIFD) generally outperform the GAN-free methods (i.e., GI, IG) on FFHQ. This indicates that the special data distribution of human faces can be more easily learned by the generative model, so that the gain from the GAN prior is larger. We also observe that the GAN-based method GGL, which only optimizes the latent code and does not fully exploit the GAN prior, yields unsatisfactory results and performs even worse than GAN-free methods on the ImageNet dataset, which again verifies the necessity of searching intermediate layers. 

We observe that GIAS performs worse when paired with BigGAN compared to StyleGAN2. One factor is the greater diversity and variability in ImageNet data. More importantly, with such a large number of parameters in BigGAN, the solution space for the GAN parameter search process becomes larger and presents a great challenge, \ie, GIAS is more susceptible to the scale of GAN. In contrast, GIFD optimizes the intermediate features and achieves faithful reconstruction on both GANs, demonstrating excellent versatility.


\begin{table*}[t]
  \centering
  \caption{Comparison of GIFD with state-of-the-art baselines on OOD data of different styles.}
    \resizebox{\linewidth}{!}{\begin{tabular}{ccllllllllllll}\toprule
    \multirow{2}[0]{*}{Dataset} & \multicolumn{1}{c}{\multirow{2}[0]{*}{Method}} & \multicolumn{4}{c}{Art Painting} & \multicolumn{4}{c}{Photo}     & \multicolumn{4}{c}{Cartoon} \\ \cmidrule(lr){3-6} \cmidrule(lr){7-10} \cmidrule(lr){11-14}
          &       & \multicolumn{1}{c}{PSNR$\uparrow$} & \multicolumn{1}{c}{LPIPS$\downarrow$} & \multicolumn{1}{c}{SSIM$\uparrow$} & \multicolumn{1}{c}{MSE$\downarrow$} & \multicolumn{1}{c}{PSNR$\uparrow$} & \multicolumn{1}{c}{LPIPS$\downarrow$} & \multicolumn{1}{c}{SSIM$\uparrow$} & \multicolumn{1}{c}{MSE$\downarrow$} & \multicolumn{1}{c}{PSNR$\uparrow$} & \multicolumn{1}{c}{LPIPS$\downarrow$} & \multicolumn{1}{c}{SSIM$\uparrow$} & \multicolumn{1}{c}{MSE$\downarrow$} \\ \midrule
    \multirow{5}[0]{*}{ImageNet*} & IG \cite{geiping2020inverting}    & 18.3476  & 0.2286  & 0.3870  & 0.0172  & 15.6647  & 0.3575  & 0.2409  & 0.0325  & 15.8766  & 0.3183  & 0.3970  & 0.0288  \\
          & GI \cite{yin2021see}    & 17.4681  & 0.2625  & 0.3445  & 0.0203  & 15.2700  & 0.3888  & 0.2201  & 0.0346  & 15.3905  & 0.3112  & 0.3926  & 0.0327  \\
          & GGL \cite{li2022auditing}   & 12.8011  & 0.3639  & 0.1356  & 0.0571  & 12.9246  & 0.3159  & 0.1507  & 0.0667  & 11.0315  & 0.3294  & 0.2832  & 0.0895  \\
          & GIAS \cite{jeon2021gradient}  & 17.2804  & 0.2774  & 0.3346  & 0.0227  & 20.4539  & 0.1724  & 0.4913  & 0.0111  & 19.0247  & 0.1862  & 0.5740  & 0.0149  \\
          & \textbf{GIFD}  & \textbf{19.3311} & \textbf{0.1700 } & \textbf{0.4503 } & \textbf{0.0151 } & \textbf{21.9281 } & \textbf{0.1137 } & \textbf{0.5765 } & \textbf{0.0082 } & \textbf{22.8055 } & \textbf{0.1030 } & \textbf{0.6970 } & \textbf{0.0067 } \\  \midrule
    \multirow{5}[0]{*}{FFHQ*} & IG \cite{geiping2020inverting}    & 15.9020  & 0.3856  & 0.2736  & 0.0273  & 17.7422  & 0.3043  & 0.3398  & 0.0174  & 14.7029  & 0.3118  & 0.3213  & 0.0358  \\
          & GI \cite{yin2021see}    & 16.2990  & 0.3537  & 0.2917  & 0.0259  & 18.5540  & 0.2388  & 0.3808  & 0.0147  & 15.0097  & 0.3232  & 0.3201  & 0.0331  \\
          & GGL \cite{li2022auditing}   & 14.2833  & 0.2514  & 0.1982  & 0.0435  & 15.5001  & 0.2309  & 0.2513  & 0.0302  & 12.3590  & 0.2556  & 0.2322  & 0.0624  \\
          & GIAS \cite{jeon2021gradient}  & 18.4619  & 0.1912  & 0.4424  & 0.0172  & 19.6763  & 0.1615  & 0.4885  & 0.0123  & 15.3798  & 0.2250  & 0.3837  & 0.0338  \\
          & \textbf{GIFD}  & \textbf{19.8847 } & \textbf{0.1534 } & \textbf{0.4979 } & \textbf{0.0120 } & \textbf{21.3981 } & \textbf{0.1148 } & \textbf{0.5446 } & \textbf{0.0098 } & \textbf{17.4005 } & \textbf{0.1634 } & \textbf{0.4614 } & \textbf{0.0220 } \\ \bottomrule
    \end{tabular}%
    }
  \label{ood_table}%
\end{table*}
\begin{figure*}[t]

    \begin{subfigure}{0.49\linewidth}
        \begin{minipage}[t]{0.03\linewidth}
        \rotatebox{90}{\scriptsize{\textbf{Art Painting}}}
        \end{minipage}%
	\begin{minipage}[t]{0.165\linewidth}
        \centering
        \includegraphics[width=1.4cm]{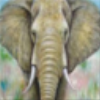}
        \end{minipage}%
        \begin{minipage}[t]{0.165\linewidth}
        \centering
        \includegraphics[width=1.4cm]{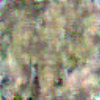}
        \end{minipage}%
        \begin{minipage}[t]{0.165\linewidth}
        \centering
        \includegraphics[width=1.4cm]{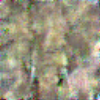}
        \end{minipage}%
        \begin{minipage}[t]{0.165\linewidth}
        \centering
        \includegraphics[width=1.4cm]{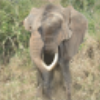}
        \end{minipage}%
        \begin{minipage}[t]{0.165\linewidth}
        \centering
        \includegraphics[width=1.4cm]{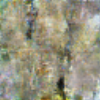}
        \end{minipage}%
        \begin{minipage}[t]{0.165\linewidth}
        \centering
        \includegraphics[width=1.4cm]{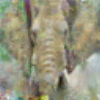}
        \end{minipage}%
        \end{subfigure}   
    \begin{subfigure}{0.49\linewidth}
        \quad
        \begin{minipage}[t]{0.165\linewidth}
        \centering
        \includegraphics[width=1.4cm]{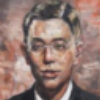}
        \end{minipage}%
        \begin{minipage}[t]{0.165\linewidth}
        \centering
        \includegraphics[width=1.4cm]{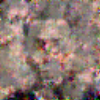}
        \end{minipage}%
        \begin{minipage}[t]{0.165\linewidth}
        \centering
        \includegraphics[width=1.4cm]{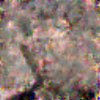}
        \end{minipage}%
        \begin{minipage}[t]{0.165\linewidth}
        \centering
        \includegraphics[width=1.4cm]{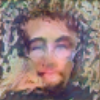}
        \end{minipage}%
        \begin{minipage}[t]{0.165\linewidth}
        \centering
        \includegraphics[width=1.4cm]{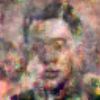}
        \end{minipage}%
        \begin{minipage}[t]{0.165\linewidth}
        \centering
        \includegraphics[width=1.4cm]{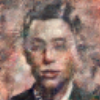}
        \end{minipage}%
        \end{subfigure}

    \begin{subfigure}{0.49\linewidth}
        \begin{minipage}[t]{0.03\linewidth}
        \rotatebox{90}{\scriptsize{\textbf{~}}}     
        \end{minipage}%
	\begin{minipage}[t]{0.165\linewidth}
        \centering
        \includegraphics[width=1.4cm]{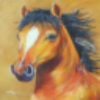}
        \end{minipage}%
        \begin{minipage}[t]{0.165\linewidth}
        \centering
        \includegraphics[width=1.4cm]{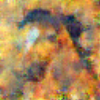}
        \end{minipage}%
        \begin{minipage}[t]{0.165\linewidth}
        \centering
        \includegraphics[width=1.4cm]{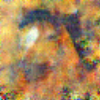}
        \end{minipage}%
        \begin{minipage}[t]{0.165\linewidth}
        \centering
        \includegraphics[width=1.4cm]{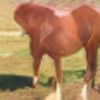}
        \end{minipage}%
        \begin{minipage}[t]{0.165\linewidth}
        \centering
        \includegraphics[width=1.4cm]{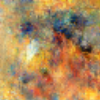}
        \end{minipage}%
        \begin{minipage}[t]{0.165\linewidth}
        \centering
        \includegraphics[width=1.4cm]{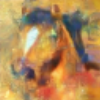}
        \end{minipage}%
        \end{subfigure}   
    \begin{subfigure}{0.49\linewidth}
        \quad
        \begin{minipage}[t]{0.165\linewidth}
        \centering
        \includegraphics[width=1.4cm]{paper_images/extra/main/extra_ffhq1.pdf}
        \end{minipage}%
        \begin{minipage}[t]{0.165\linewidth}
        \centering
        \includegraphics[width=1.4cm]{paper_images/extra/main/extra_ffhq2.pdf}
        \end{minipage}%
        \begin{minipage}[t]{0.165\linewidth}
        \centering
        \includegraphics[width=1.4cm]{paper_images/extra/main/extra_ffhq3.pdf}
        \end{minipage}%
        \begin{minipage}[t]{0.165\linewidth}
        \centering
        \includegraphics[width=1.4cm]{paper_images/extra/main/extra_ffhq4.pdf}
        \end{minipage}%
        \begin{minipage}[t]{0.165\linewidth}
        \centering
        \includegraphics[width=1.4cm]{paper_images/extra/main/extra_ffhq5.pdf}
        \end{minipage}%
        \begin{minipage}[t]{0.165\linewidth}
        \centering
        \includegraphics[width=1.4cm]{paper_images/extra/main/extra_ffhq6.pdf}
        \end{minipage}%
        \end{subfigure}

    \begin{subfigure}{0.49\linewidth}
        \begin{minipage}[t]{0.03\linewidth}
        \rotatebox{90}{\scriptsize{\textbf{~~~~~~~~Photo}}}         
        \end{minipage}%
        \begin{minipage}[t]{0.165\linewidth}
        \centering
        \includegraphics[width=1.4cm]{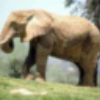}
        \end{minipage}%
        \begin{minipage}[t]{0.165\linewidth}
        \centering
        \includegraphics[width=1.4cm]{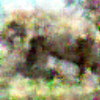}
        \end{minipage}%
        \begin{minipage}[t]{0.165\linewidth}
        \centering
        \includegraphics[width=1.4cm]{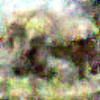}
        \end{minipage}%
        \begin{minipage}[t]{0.165\linewidth}
        \centering
        \includegraphics[width=1.4cm]{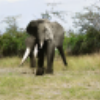}
        \end{minipage}%
        \begin{minipage}[t]{0.165\linewidth}
        \centering
        \includegraphics[width=1.4cm]{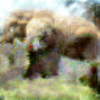}
        \end{minipage}%
        \begin{minipage}[t]{0.165\linewidth}
        \centering
        \includegraphics[width=1.4cm]{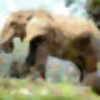}
        \end{minipage}%
        \end{subfigure}   
    \begin{subfigure}{0.49\linewidth}
        \quad
        \begin{minipage}[t]{0.165\linewidth}
        \centering
        \includegraphics[width=1.4cm]{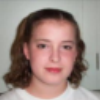}
        \end{minipage}%
        \begin{minipage}[t]{0.165\linewidth}
        \centering
        \includegraphics[width=1.4cm]{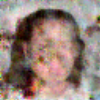}
        \end{minipage}%
        \begin{minipage}[t]{0.165\linewidth}
        \centering
        \includegraphics[width=1.4cm]{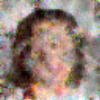}
        \end{minipage}%
        \begin{minipage}[t]{0.165\linewidth}
        \centering
        \includegraphics[width=1.4cm]{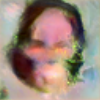}
        \end{minipage}%
        \begin{minipage}[t]{0.165\linewidth}
        \centering
        \includegraphics[width=1.4cm]{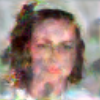}
        \end{minipage}%
        \begin{minipage}[t]{0.165\linewidth}
        \centering
        \includegraphics[width=1.4cm]{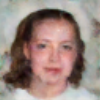}
        \end{minipage}%
        \end{subfigure}

    \begin{subfigure}{0.49\linewidth}
        \begin{minipage}[t]{0.03\linewidth}
        \rotatebox{90}{\scriptsize{\textbf{}}}     
        \end{minipage}%
	\begin{minipage}[t]{0.165\linewidth}
        \centering
        \includegraphics[width=1.4cm]{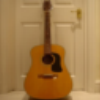}
        \end{minipage}%
        \begin{minipage}[t]{0.165\linewidth}
        \centering
        \includegraphics[width=1.4cm]{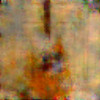}
        \end{minipage}%
        \begin{minipage}[t]{0.165\linewidth}
        \centering
        \includegraphics[width=1.4cm]{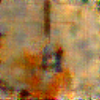}
        \end{minipage}%
        \begin{minipage}[t]{0.165\linewidth}
        \centering
        \includegraphics[width=1.4cm]{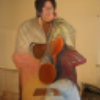}
        \end{minipage}%
        \begin{minipage}[t]{0.165\linewidth}
        \centering
        \includegraphics[width=1.4cm]{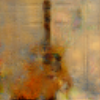}
        \end{minipage}%
        \begin{minipage}[t]{0.165\linewidth}
        \centering
        \includegraphics[width=1.4cm]{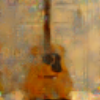}
        \end{minipage}%
        \end{subfigure}   
    \begin{subfigure}{0.49\linewidth}
        \quad
        \begin{minipage}[t]{0.165\linewidth}
        \centering
        \includegraphics[width=1.4cm]{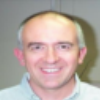}
        \end{minipage}%
        \begin{minipage}[t]{0.165\linewidth}
        \centering
        \includegraphics[width=1.4cm]{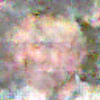}
        \end{minipage}%
        \begin{minipage}[t]{0.165\linewidth}
        \centering
        \includegraphics[width=1.4cm]{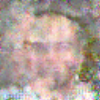}
        \end{minipage}%
        \begin{minipage}[t]{0.165\linewidth}
        \centering
        \includegraphics[width=1.4cm]{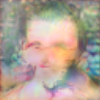}
        \end{minipage}%
        \begin{minipage}[t]{0.165\linewidth}
        \centering
        \includegraphics[width=1.4cm]{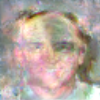}
        \end{minipage}%
        \begin{minipage}[t]{0.165\linewidth}
        \centering
        \includegraphics[width=1.4cm]{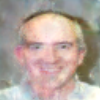}
        \end{minipage}%
        \end{subfigure}

    \begin{subfigure}{0.49\linewidth}
        \begin{minipage}[t]{0.03\linewidth}
        \rotatebox{90}{\scriptsize{\textbf{~~~~~Cartoon}}}     
        \end{minipage}%
	\begin{minipage}[t]{0.165\linewidth}
        \centering
        \includegraphics[width=1.4cm]{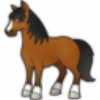}
        \end{minipage}%
        \begin{minipage}[t]{0.165\linewidth}
        \centering
        \includegraphics[width=1.4cm]{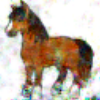}
        \end{minipage}%
        \begin{minipage}[t]{0.165\linewidth}
        \centering
        \includegraphics[width=1.4cm]{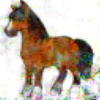}
        \end{minipage}%
        \begin{minipage}[t]{0.165\linewidth}
        \centering
        \includegraphics[width=1.4cm]{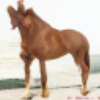}
        \end{minipage}%
        \begin{minipage}[t]{0.165\linewidth}
        \centering
        \includegraphics[width=1.4cm]{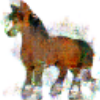}
        \end{minipage}%
        \begin{minipage}[t]{0.165\linewidth}
        \centering
        \includegraphics[width=1.4cm]{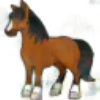}
        \end{minipage}%
        \end{subfigure}   
    \begin{subfigure}{0.49\linewidth}
        \quad
        \begin{minipage}[t]{0.165\linewidth}
        \centering
        \includegraphics[width=1.4cm]{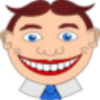}
        \end{minipage}%
        \begin{minipage}[t]{0.165\linewidth}
        \centering
        \includegraphics[width=1.4cm]{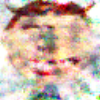}
        \end{minipage}%
        \begin{minipage}[t]{0.165\linewidth}
        \centering
        \includegraphics[width=1.4cm]{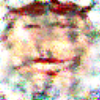}
        \end{minipage}%
        \begin{minipage}[t]{0.165\linewidth}
        \centering
        \includegraphics[width=1.4cm]{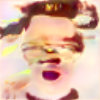}
        \end{minipage}%
        \begin{minipage}[t]{0.165\linewidth}
        \centering
        \includegraphics[width=1.4cm]{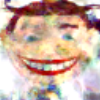}
        \end{minipage}%
        \begin{minipage}[t]{0.165\linewidth}
        \centering
        \includegraphics[width=1.4cm]{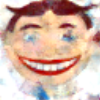}
        \end{minipage}%
        \end{subfigure}

    \begin{subfigure}{0.49\linewidth}
        \begin{minipage}[t]{0.03\linewidth}
        \rotatebox{90}{\scriptsize{}}     
        \end{minipage}%
        \begin{minipage}[t]{0.165\linewidth}
        \centering
        \includegraphics[width=1.4cm]{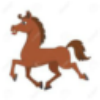}
        \centering
        \caption*{\textbf{\footnotesize{Original}}}
        \end{minipage}%
        \begin{minipage}[t]{0.165\linewidth}
        \centering
        \includegraphics[width=1.4cm]{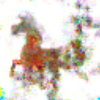}
        \centering
        \caption*{\textbf{\footnotesize{{}IG \cite{geiping2020inverting}}}}
        \end{minipage}%
        \begin{minipage}[t]{0.165\linewidth}
        \centering
        \includegraphics[width=1.4cm]{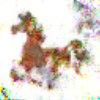}
        \centering
        \caption*{\textbf{\footnotesize{GI \cite{yin2021see}}}}
        \end{minipage}%
        \begin{minipage}[t]{0.165\linewidth}
        \centering
        \includegraphics[width=1.4cm]{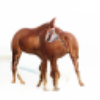}
        \centering
        \caption*{\textbf{\footnotesize{GGL \cite{li2022auditing}}}}
        \end{minipage}%
        \begin{minipage}[t]{0.165\linewidth}
        \centering
        \includegraphics[width=1.4cm]{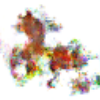}
        \centering
        \caption*{\textbf{\footnotesize{GIAS \cite{jeon2021gradient}}}}
        \end{minipage}%
        \begin{minipage}[t]{0.165\linewidth}
        \centering
        \includegraphics[width=1.4cm]{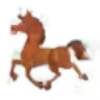}
        \centering
        \caption*{\textbf{\footnotesize{GIFD}}}
        \end{minipage}
	  \caption{ImageNet* (BigGAN)}
        \end{subfigure}
    \begin{subfigure}{0.49\linewidth}
    \quad
        \begin{minipage}[t]{0.165\linewidth}
        \centering
        \includegraphics[width=1.4cm]{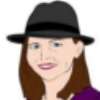}
        \centering
        \caption*{\textbf{\footnotesize{Original}}}
        \end{minipage}%
        \begin{minipage}[t]{0.165\linewidth}
        \centering
        \includegraphics[width=1.4cm]{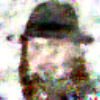}
        \centering
        \caption*{\textbf{\footnotesize{IG \cite{geiping2020inverting}}}}
        \end{minipage}%
        \begin{minipage}[t]{0.165\linewidth}
        \centering
        \includegraphics[width=1.4cm]{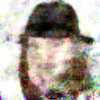}
        \centering
        \caption*{\textbf{\footnotesize{GI \cite{yin2021see}}}}
        \end{minipage}%
        \begin{minipage}[t]{0.165\linewidth}
        \centering
        \includegraphics[width=1.4cm]{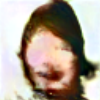}
        \centering
        \caption*{\textbf{\footnotesize{GGL \cite{li2022auditing}}}}
        \end{minipage}%
        \begin{minipage}[t]{0.165\linewidth}
        \centering
        \includegraphics[width=1.4cm]{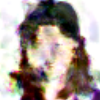}
        \centering
        \caption*{\textbf{\footnotesize{GIAS \cite{jeon2021gradient}}}}
        \end{minipage}%
        \begin{minipage}[t]{0.165\linewidth}

        \includegraphics[width=1.4cm]{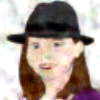}
        \centering
        \caption*{\footnotesize{\textbf{GIFD}}}
        \end{minipage}
	
        \caption{FFHQ* (StyleGAN2)}
        \end{subfigure}

    \caption{Visual comparison of different methods on ImageNet* and FFHQ*. }
    \label{ood_figure}
\end{figure*}

\subsection{Out of Distribution Data Recovery}
We then consider a more practical scenario where the training sets of the GAN model and the FL task obey different data distributions. Note that the scope of our analysis is the generalization of GIFD to unseen image data under the considered OOD settings, including style shift and label inconsistency. We first consider the OOD data with the same label space, but quite different feature distributions. Hereinafter, we denote the OOD data of ImageNet and FFHQ by ImageNet* and FFHQ*, respectively.

PACS \cite{li2017deeper} is a widely used benchmark for domain generalization with four different styles, \ie, Art Painting, Cartoon, Photo, and Sketch. To achieve our OOD setting, we manually select data with three different styles (\ie, Art Painting, Cartoon, Photo) from the validation set. For each style in ImageNet*, we select 15 images of guitar, elephant, and horse in total. For FFHQ*, we select 15 images for each style and crop them to obtain the face images. We present visual comparison and quantitative results in Figure \ref{ood_figure} and Table \ref{ood_table}. 
Besides, we further consider more challenging scenarios where we use GANs to attack datasets with completely non-repetitive category space, \ie, BigGAN pre-trained on ImageNet for inverting FFHQ, as shown in Table \ref{tab:label_mapping}. Since BigGAN is a label-conditioned model, we can use the proposed label mapping technique to enhance the target label information.

\noindent\textbf{Experiment Results.} Experimental results in Table \ref{ood_table} demonstrate the significant improvement of GIFD over the baselines. GAN-based methods perform best on Photo, whose domain characteristics are similar to their training data. However, on Art in ImageNet*, most GAN-based methods underperform even GAN-free baselines, indicating that their limited output diversity hinders reconstruction across different styles. In contrast, GIFD enlarges the output space by better exploiting GAN priors, enabling superior performance and therefore relaxing the assumption that the generative model and FL data must follow the same feature distribution.

\begin{table}[t]
    \centering
    \caption{Performance comparison on ImageNet using StyleGAN2 trained exclusively on human faces (FFHQ).}
    \label{tab:ood_extreme}
    \resizebox{0.95\linewidth}{!}{
    \begin{tabular}{lcccc}
        \toprule
        Method & PSNR~$\uparrow$ & LPIPS~$\downarrow$ & SSIM~$\uparrow$ & MSE~$\downarrow$ \\
        \midrule
        IG~\cite{geiping2020inverting} 
        & 17.0756 & 0.3078 & 0.2908 & 0.0223 \\
        
        GI~\cite{yin2021see} 
        & 16.5109 & 0.3297 & 0.2673 & 0.0258 \\
        
        GGL~\cite{li2022auditing} 
        & 14.2475 & 0.3705 & 0.1546 & 0.0458 \\
        
        GIAS~\cite{jeon2021gradient} 
        & 18.1455 & 0.2298 & 0.3826 & 0.0207 \\
        
        \textbf{GIFD}
        & \textbf{19.3660} & \textbf{0.1997} & \textbf{0.4113} & \textbf{0.0152} \\
        \bottomrule
    \end{tabular}
    }
\end{table}

To investigate the performance under an extreme mismatch between the generative prior and the private data, we further conduct an experiment using a StyleGAN2 model trained exclusively on human faces to reconstruct diverse images from ImageNet, where the GAN training domain has almost no semantic overlap with the target ImageNet classes. As in Table \ref{tab:ood_extreme}, many attacks exhibit degraded performance compared with using the ImageNet-trained BigGAN. Nevertheless, GIFD consistently outperforms baselines and achieves satisfactory results, again demonstrating its ability to generalize to recover privacy even with severe semantic mismatch.

\begin{table}[t]
  \centering
  \caption{Comparison on more challenging OOD scenarios of label inconsistency. We attack the FL model trained on FFHQ using the GAN trained on ImageNet. Note that GIFD* denotes GIFD with a label-mapping technique.}
    \resizebox{0.89\linewidth}{!}{\begin{tabular}{lcccc} \toprule Method & \multicolumn{1}{l}{PSNR$\uparrow$} & \multicolumn{1}{l}{LPIPS$\downarrow$} & \multicolumn{1}{l}{SSIM$\uparrow$} & \multicolumn{1}{l}{MSE$\downarrow$} \\ \midrule IG \cite{geiping2020inverting} & 15.3523 & 0.4172 & 0.2272 & 0.0311 \\ GI \cite{yin2021see} & 14.9485 & 0.4503 & 0.2044 & 0.0343 \\ GGL \cite{li2022auditing} & 11.1235 & 0.4498 & 0.0995 & 0.0843 \\ GIAS \cite{jeon2021gradient} & 17.4565 & 0.2534 & 0.3825 & 0.0201 \\ \textbf{GIFD} & 19.9793 & 0.1597 & 0.5095 & 0.0126 \\ \textbf{GIFD*} & \textbf{20.3664} & \textbf{0.1463} & \textbf{0.5288} & \textbf{0.0114} \\ \bottomrule \end{tabular}
}
  \label{tab:label_mapping}%
\end{table}%

Moreover, we present results under the more challenging scenario of label inconsistency in Table \ref{tab:label_mapping}, \ie, the label space of the GAN's training set has no overlap with the private data. These results again confirm the superiority of GIFD under practical OOD settings. Notably, the attack performance of most existing methods decreases compared with the former in-distribution settings due to the larger distribution gap, yet our GIFD still achieves satisfactory results, highlighting that the label mapping mechanism further brings notable improvements by rectifying label information.


\begin{table}[htbp]

  \caption{PSNR mean of different attack methods under different defense strategies.}
    \begin{subtable}[t]{\linewidth}
    \resizebox{\linewidth}{!}{\begin{tabular}{lccccc}
    \toprule
    \multicolumn{1}{l}{\multirow{2}[0]{*}{Defenses}} & \multicolumn{5}{c}{Method} \\ \cmidrule(lr){2-6}
    & IG \cite{geiping2020inverting} & GI \cite{yin2021see} & GGL \cite{li2022auditing} & GIAS \cite{jeon2021gradient} & \textbf{GIFD} \\ \midrule
           Noise   & 11.0654  & 10.0818 & 12.7640  & 12.5397 & \textbf{13.2558} \\
           Clipping    & 16.4418  & 12.5387  & 12.7930  & 17.9384 & \textbf{18.8983} \\
           Sparsification   & 12.0760  & 12.1691  & 12.6810  & 15.1745 & \textbf{16.0240} \\
           Soteria  & 9.1941  & 10.1831  & 12.8433  & 16.8151 & \textbf{18.3205} \\
           ATSPrivacy  & 19.1398 & 17.4004 & 12.4586 & 18.1921 & \textbf{20.5383} \\
           CENSOR & 13.7162 & 13.8353 & 13.8557 & 13.3060 & \textbf{14.6617} \\ \bottomrule
\end{tabular}}
    
    \vspace{0.05pt}
    \caption{ImageNet}
    \end{subtable}
    
    \begin{subtable}[t]{\linewidth}
    \resizebox{\linewidth}{!}{\begin{tabular}{lccccc}
    \toprule
    \multicolumn{1}{l}{\multirow{2}[0]{*}{Defenses}} & \multicolumn{5}{c}{Method} \\ \cmidrule(lr){2-6}
    & IG \cite{geiping2020inverting} & GI \cite{yin2021see} & GGL \cite{li2022auditing} & GIAS \cite{jeon2021gradient} & \textbf{GIFD} \\ \midrule
           Noise   & 11.2766  & 10.4968 & \textbf{14.8982}  & 12.1276 & 13.7118 \\
           Clipping    & 18.1382  & 12.4146  & 15.6669  & 20.4726 & \textbf{21.2861} \\
           Sparsification   & 12.0077  & 12.1849  & 14.9123  & 16.7005 & \textbf{17.3253} \\
           Soteria  & 9.8334  & 10.0843  & 15.1798  & 20.4283 & \textbf{21.1545} \\
           ATSPrivacy  & 18.1728 & 16.9409  & 15.0417 & 19.1092 & \textbf{19.4446} \\
           CENSOR & 9.3666 & 9.3427 & 9.7257 & 9.7430 & \textbf{10.7994} \\ \bottomrule
\end{tabular}}
    
    \vspace{0.05pt}
    \caption{FFHQ}

    \end{subtable}

  \label{defense_table}%
    
\end{table}%
\subsection{Attacks under Certain Defense Strategies}

Next, we consider a more robust and secure FL system with defense strategies. To make a fair comparison, we equip all the baselines with the gradient transformation technique mentioned earlier to mitigate the defense impact.

\begin{figure}[t]
    \centering
    \includegraphics[width=\linewidth]{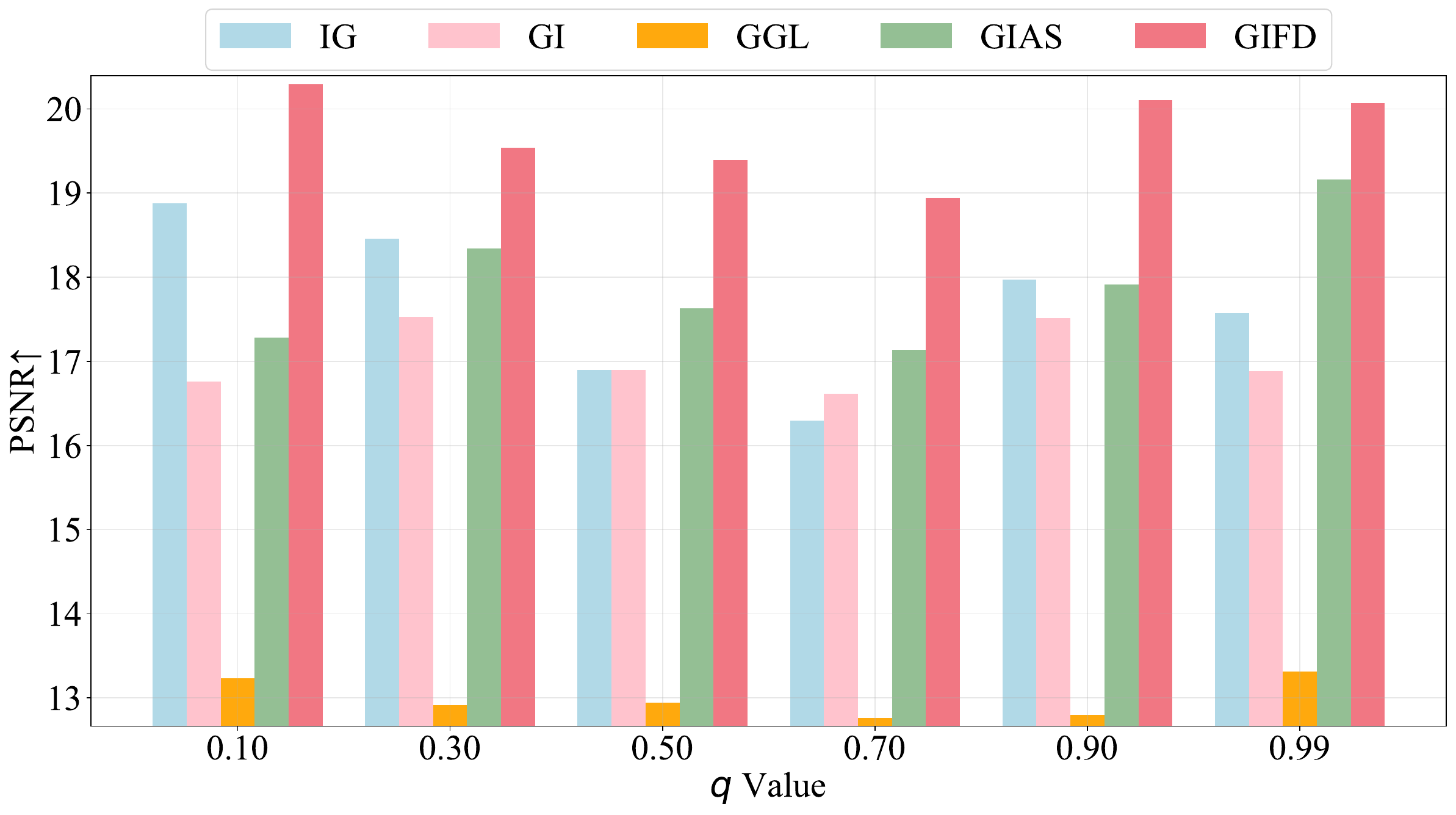}
    \caption{PSNR results of different attacks when changing the values of $q$ (the probability of assigning a training example with a certain label to a particular client) on ImageNet. These settings strictly align with the previous works \cite{wei2025extracting, cao2021fltrust} considering data heterogeneity.}
    \label{more_q_value}
\end{figure}

Following the previous works \cite{li2022auditing, gao2023automatic}, we consider the strict defense settings: (1) \textit{Gaussian Noise} with standard deviation 0.1; (2) \textit{Gradient Clipping} with a clip bound of 4; (3) \textit{Gradient Sparsification} in a sparsity of 90\%; (4) \textit{Soteria} with a pruning rate of 80\% \cite{sun2021soteria}; (5) \textit{ATSPrivacy} with a data augmentation library consisting of 1500 policies \cite{gao2023automatic}; and (6) \textit{CENSOR} \cite{zhang2025censor} by perturb gradients within a subspace orthogonal to the original gradient.

\noindent\textbf{Experiment Results.} We present experiment results in Table \ref{defense_table} compared to related methods. With the underlying gradient transformation and the fully exploited GAN image prior, GIFD is still able to invert a degraded gradient observation to reveal private information, especially in cases of clipping and Soteria. One exception is that GGL performs best on FFHQ under additive noise. This is because the high-variance Gaussian noise severely corrupts the gradients, making high-quality reconstruction difficult, while GGL can still recover coarse facial contours through latent-space optimization, yielding seemingly competitive metrics despite poor visual fidelity. This result further suggests that Gaussian noise can effectively defend against GLA when its variance is sufficiently large. Moreover, CENSOR, a recent powerful method specifically designed to defend against gradient inversion attacks, also exhibits outstanding defense effectiveness. However, our method still exhibits better resilience than baselines.

\subsection{Attacks under Different FL Settings}

\textbf{Attack Results on Heterogeneous Data.} Data heterogeneity is a well-recognized challenge in FL systems \cite{ye2023heterogeneous}. To further evaluate the robustness of GIFD, we follow the settings in prior works \cite{wei2025extracting, cao2021fltrust} and conduct heterogeneous-data experiments. Concretely, we randomly distribute dataset classes among $N_{\text{client}}$ clients: a sample with label $\ell$ is assigned to a specific client with probability $q$, and to any other client with probability $\frac{1-q}{N_{\text{client}}-1}$. When $q = \frac{1}{N_{\text{client}}}$, the local datasets follow an IID distribution; otherwise, they are non-IID. We set $N_{\text{client}} = 10$, so the IID case corresponds to $q=\frac{1}{10}$. We then test all attack approaches under $q \in \{0.10, 0.30, 0.50, 0.70, 0.90, 0.99\}$.

As in Figure \ref{more_q_value}, GIFD consistently achieves the best performance across all $q$ values. This demonstrates that GIFD remains highly effective under varying degrees of data heterogeneity, underscoring its adaptability in practical FL scenarios.

\begin{figure*}[t]
    \centering
    \begin{subfigure}{0.2455\linewidth}
        \includegraphics[width=\linewidth]{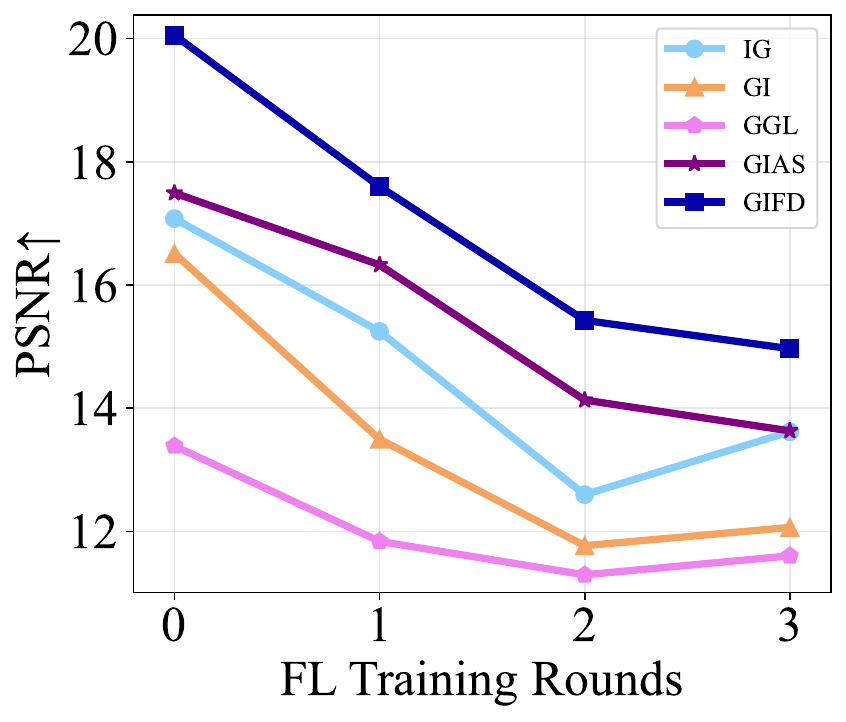}
    \end{subfigure}
    \begin{subfigure}{0.2455\linewidth}
        \includegraphics[width=\linewidth]{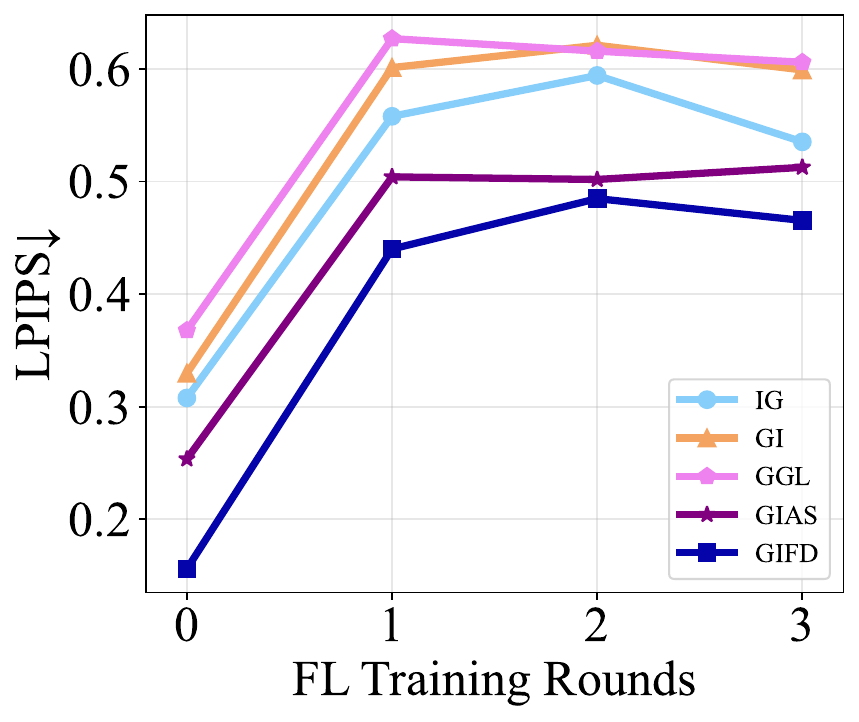}
    \end{subfigure}
    \begin{subfigure}{0.2455\linewidth}
        \includegraphics[width=\linewidth]{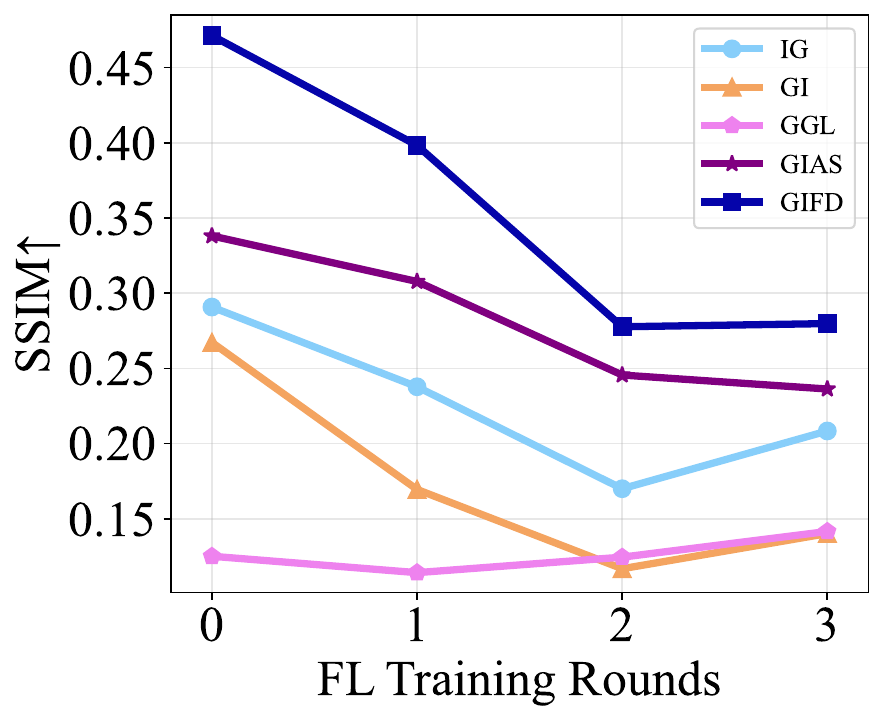}
    \end{subfigure}
    \begin{subfigure}{0.2455\linewidth}
        \includegraphics[width=\linewidth]{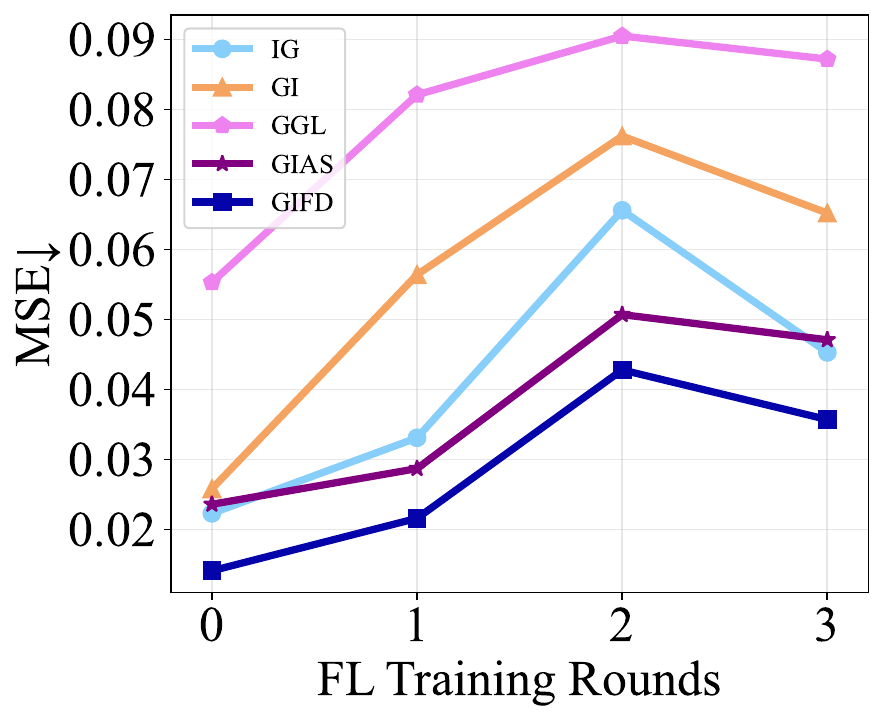}
    \end{subfigure}

    \caption{Comparison of GIFD with state-of-the-art baselines on ImageNet when increasing the FL training rounds.}
    \label{more_fl_training_rounds}
\end{figure*}

\textbf{Attack Results on More FL training rounds.} Most existing gradient inversion methods \cite{li2022auditing, jeon2021gradient} evaluate attack performance only at the very early stage of FL training, where the global model is assumed to be randomly initialized. To assess the robustness in a more practical attack scenario, we gradually increase the number of FL training rounds and accordingly perform GLA on the corresponding global model. As shown in Figure \ref{more_fl_training_rounds}, the attack performance of all methods degrades as the number of rounds increases. This is likely because, as the model converges, its gradients become smoother and less input-specific, carrying weaker signals that make reconstruction increasingly difficult. Nevertheless, GIFD consistently outperforms all baselines, demonstrating its robustness and superiority in more realistic FL scenarios.

\subsection{Further Analysis}

\textbf{Performance of Larger Batch Sizes.} We then increase the batch size and observe the results of each algorithm. Notably, we assume that no duplicate labels in each batch and infer the labels from the received gradients \cite{yin2021see}. We present the results on ImageNet and FFHQ in Table \ref{bs}. Note that the latent vector of StyleGAN2 contains a relatively large number of parameters to be optimized, while the CMA-ES optimizer adopted by GGL does not support high-dimensional optimization. Therefore, we experimentally find that GGL becomes infeasible to run when $B>2$.

\begin{table}[htbp]
  \centering
  \caption{PSNR mean of different attack methods under different batch sizes.}
    \begin{subtable}[t]{\linewidth}
    \resizebox{\linewidth}{!}{\begin{tabular}{lrrrrrr} \toprule
    \multicolumn{1}{l}{\multirow{2}[0]{*}{Method}} & \multicolumn{6}{c}{Batch Size} \\ \cmidrule(lr){2-7}
          & \multicolumn{1}{c}{1} & \multicolumn{1}{c}{2} & \multicolumn{1}{c}{4} & \multicolumn{1}{c}{8} & \multicolumn{1}{c}{16} & \multicolumn{1}{c}{32} \\ \midrule
    IG \cite{geiping2020inverting}   & 17.4634  & 15.2417  & 14.3744  & 13.6599  & 13.1545  & 12.0795  \\
    GI \cite{yin2021see}   & 17.4373  & 14.7293  & 14.0947  & 13.3001  & 12.7842  & 11.8767  \\
    GGL \cite{li2022auditing}  & 12.7511  & 12.8903  & 13.1875  & 12.6001  & 11.8027  & 11.0896  \\
    GIAS \cite{jeon2021gradient} & 17.1401  & 16.1683  & 15.5894  & 15.2130  & 14.4462  & 13.6080  \\
    \textbf{GIFD}  & \textbf{20.6217}  & \textbf{16.7542} & \textbf{16.4272}  & \textbf{15.4889}  & \textbf{14.6500}  & \textbf{13.8106}  \\
    \bottomrule
\end{tabular}

    }
    \vspace{0.05pt}
    \caption{ImageNet}
    \end{subtable}
    
    \begin{subtable}[t]{\linewidth}
    \centering
    \resizebox{\linewidth}{!}{\begin{tabular}{lcccccc} \toprule
    \multicolumn{1}{l}{\multirow{2}[0]{*}{Method}} & \multicolumn{6}{c}{Batch Size} \\ \cmidrule(lr){2-7}
          & \multicolumn{1}{c}{1} & \multicolumn{1}{c}{2} & \multicolumn{1}{c}{4} & \multicolumn{1}{c}{8} & \multicolumn{1}{c}{16} & \multicolumn{1}{c}{32} \\ \midrule
    IG \cite{geiping2020inverting}    & 19.0761 & 16.2766 & 13.8748 & 12.2449 & 10.3703 & 9.9381 \\
    GI \cite{yin2021see}    & 17.3506 & 15.5546 & 13.4236 & 12.1993 & 9.9744 & 9.9056 \\
    GGL \cite{li2022auditing}  & 14.7479 & 13.3547 &   ——   &  —— & —— & —— \\
    GIAS \cite{jeon2021gradient} & 20.0779 & 16.9557 & 13.6716 & 12.4989 & 10.5437 & 10.0206 \\
    \textbf{GIFD}  & \textbf{21.1334} & \textbf{17.9619} & \textbf{14.3493} & \textbf{12.7402} & \textbf{10.6833} & \textbf{10.2697} \\
 \bottomrule
    \end{tabular}%
    }
    \vspace{0.1pt}
    \caption{FFHQ}
    \end{subtable}
  \label{bs}%
\end{table}%

As shown in Table \ref{bs}, the proposed GIFD achieves a steady improvement over previous methods at all batch sizes. The numerical results also show that the reconstruction at large batch sizes is still a significant challenge.

\textbf{Performance on More Datasets.} To further evaluate the generalizability of our attack, we additionally consider five widely tested datasets in gradient inversion attacks \cite{zhu2019deep, geiping2020inverting, jeon2021gradient}, including CIFAR-10/CIFAR-100 \cite{krizhevsky2009learning}, MNIST \cite{deng2012mnist}, KMNIST \cite{clanuwat2018deep}, and SVHN \cite{netzer2011reading}. As shown in Table \ref{tab:datasets}, all attack methods except GGL achieve strong reconstruction performance on these datasets, while our GIFD consistently outperforms these baselines. This is largely because the images in these datasets have lower resolutions and thus lead to fewer parameters to recover, which is a relatively easier optimization problem. 
In contrast, GGL reaches limited reconstruction quality due to its restriction to latent space, further highlighting the necessity of the proposed intermediate-layer search.

\begin{table}[t]
  \centering
  \caption{PSNR mean of different attack methods across different private datasets.}
  \setlength{\tabcolsep}{4pt}
  \resizebox{\linewidth}{!}{\begin{tabular}{lccccc} \toprule
    Method & CIFAR-10 & CIFAR-100 & MNIST & KMNIST & SVHN \\ \midrule
    IG \cite{geiping2020inverting}   & 25.9120 & 21.0027 & 26.4194 & 26.3165 & 24.8600 \\
    GI \cite{yin2021see}  & 26.1532 & 22.9966 & 25.7285 & 27.1744 & 24.4489 \\
    GGL \cite{li2022auditing} & 12.5193 & 11.9671 & 10.0464 & 8.6165  & 17.8009 \\
    GIAS \cite{jeon2021gradient} & 27.5338 & 26.4250 & 39.5232 & 28.5140 & 28.3148 \\
    \textbf{GIFD} & \textbf{28.9481} & \textbf{28.6634} & \textbf{40.4472} & \textbf{34.5892} & \textbf{30.6602} \\ \bottomrule
  \end{tabular}}
  \label{tab:datasets}
\end{table}

\begin{figure}[t]
	\centering
	\begin{subfigure}{0.4935\linewidth}
		\centering
		\includegraphics[width=\linewidth]{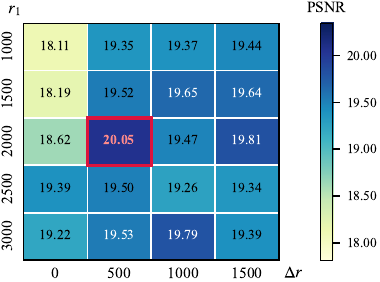}
		\caption{ImageNet (BigGAN)}
		\label{radius_imagenet}
	\end{subfigure}
	\begin{subfigure}{0.4935\linewidth}
		\centering
		\includegraphics[width=\linewidth]{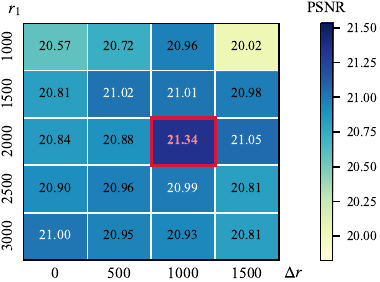}
		\caption{FFHQ (StyleGAN2)}
		\label{radius_ffhq}
	\end{subfigure}
	\centering
	\caption{PSNR of the grid search over the starting radius $r_1$ (y-axis) and the step size $\Delta r$ (x-axis) of the $l_1$ ball. The red box marks the configuration adopted in our main experiments.}
	\label{radius_search}
\end{figure}

\textbf{The choice of the radius of the $l_1$ ball.} Guided by the theory in \cite{daras2021intermediate} that a sequence of increasing radii of the $l_1$ ball tends to provide better results, we gradually increase the radius toward deeper layers to encourage exploration of a broader feature space. 
In practice, we adopt a linear growth strategy, and the radius $r_k$ of the $k$-th intermediate layer can be formulated as:

\begin{equation}
\begin{aligned}
r_k = r_1 + (k-1)\Delta r,
\label{eq:radii}
\end{aligned}
\end{equation}

\noindent where $r_1$ is the radius of the first layer and $\Delta r$ is the layer-wise step size.
We then conduct ablation studies on BigGAN and StyleGAN2 with varying values of the starting radius $r_1$ and the step size $\Delta r$. 
As shown in Figure \ref{radius_search}, $r_1=2000$ and $\Delta r=500$ reach the best results for BigGAN, while $r_1=2000$ and $\Delta r=1000$ reach the best results for StyleGAN2. 
When applying GIFD to new architectures, researchers can refer to the radius schedule in Eq. (\ref{eq:radii}) as a practical guideline and conduct ablation studies to determine the optimal radius configuration. Empirically, for GAN models with more intermediate layers, a smaller step size (\eg, 500) is recommended, while a larger step size (\eg, 1000) can be used for models with fewer intermediate layers.

\textbf{Performance on More FL Global Models.} To further validate our method, we provide numerical results of PSNR on more FL global models. In addition to various CNN architectures, we also consider transformer-based models, including variants of ViT \cite{dosovitskiy2020image} and DeiT \cite{touvron2021training}.

\begin{figure}[t]
	\centering
	\begin{subfigure}{0.88\linewidth}
		\centering
		\includegraphics[width=\linewidth]{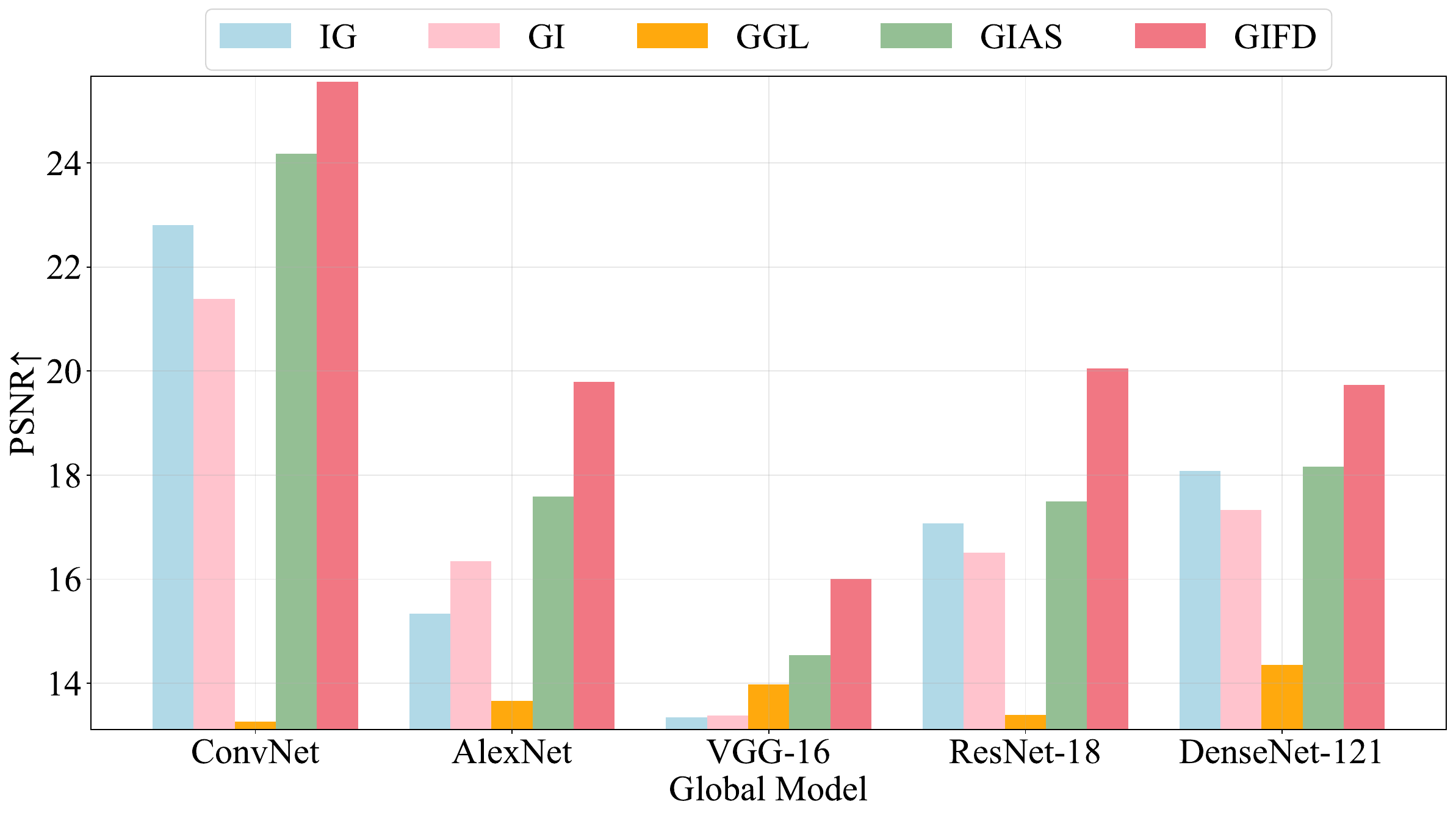}
		\vspace{-4ex} 
		\caption{CNN Architectures}
		\label{more_model_cnn}
	\end{subfigure}
	
	\vspace{1.5ex} 
	
	\begin{subfigure}{0.88\linewidth}
		\centering
		\includegraphics[width=\linewidth]{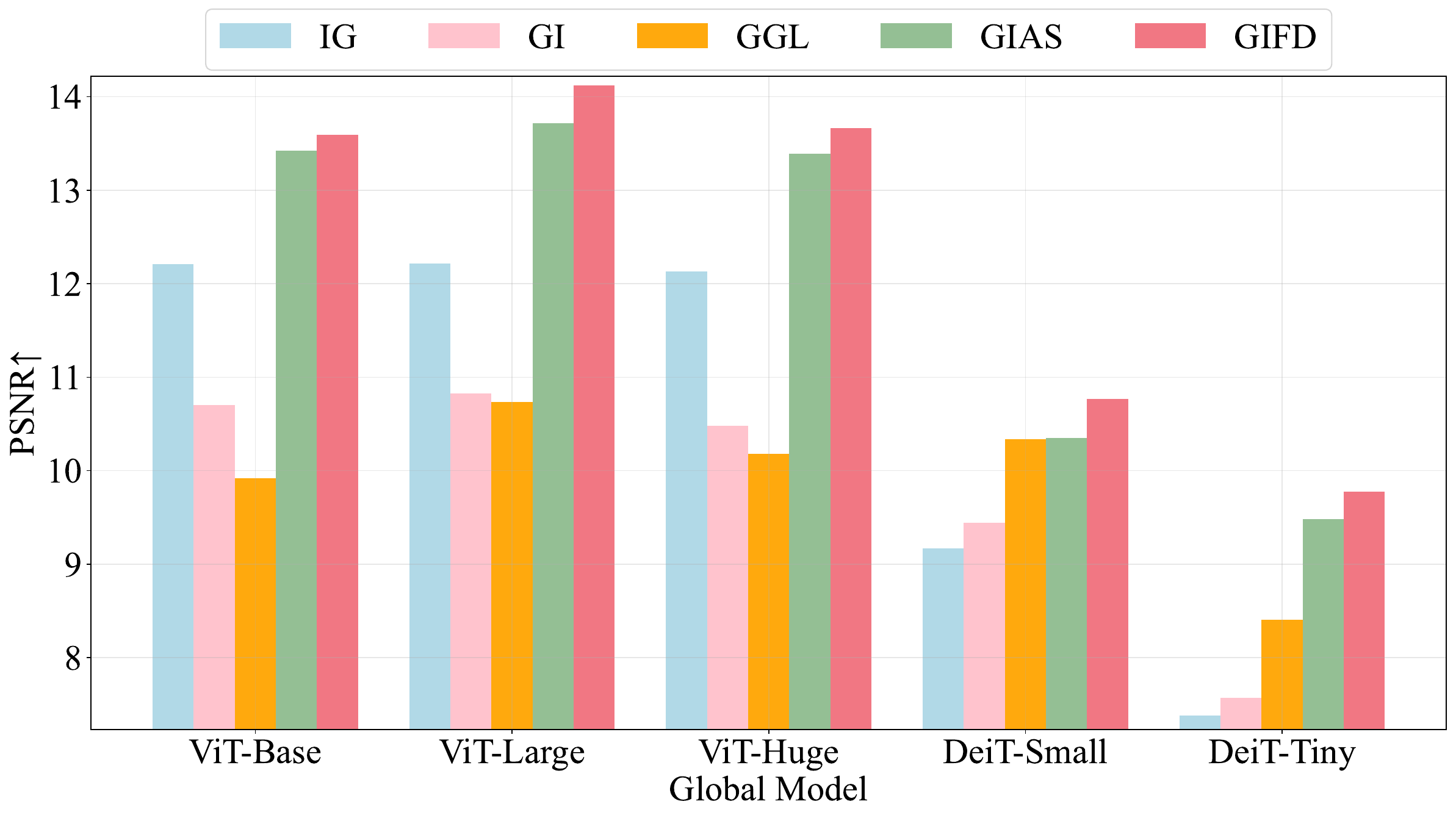}
		\vspace{-4ex} 
		\caption{Transformer Architectures}
		\label{more_model_transformer}
	\end{subfigure}
	
	\caption{PSNR results of different methods for different global models on the ImageNet dataset.}
	\label{more_model}
\end{figure}

Figure \ref{more_model} shows that the overall attack performance varies across different FL models, and GIFD always performs the best. It also suggests that the model architecture is highly related to the defense effectiveness. \textit{I.e.}, ViT-based architectures are generally more difficult to invert than CNN-based models. This may be attributed to the stronger local spatial correlations preserved by CNN gradients, whereas ViTs tokenize images into patches and mix information globally through self-attention, making the correspondence between gradients and local pixel structures less direct. This is consistent with existing findings that transformer representations exhibit stronger global feature mixing and weaker locality than convolutional representations \cite{li2023localvit}, which can reduce the amount of directly recoverable spatial information from gradients.

\begin{table}[t]
  \centering
  \caption{Ablation study of GIFD and its three variants on every 1000th image of ImageNet and FFHQ validation set.}
  
  \centering
    \begin{subtable}[t]{\linewidth}
    \centering
     \resizebox{0.8\linewidth}{!}{
     \begin{tabular}{lccccc} \toprule
    \multicolumn{2}{l}{\multirow{2}[1]{*}{Method}} & \multicolumn{4}{c}{Metric} \\ \cmidrule(lr){3-6}
     & & \multicolumn{1}{c}{PSNR$\uparrow$} & \multicolumn{1}{c}{LPIPS$\downarrow$} & \multicolumn{1}{c}{SSIM$\uparrow$} & \multicolumn{1}{c}{MSE$\downarrow$} \\  \midrule
    \multicolumn{2}{l}{GIFD-$z$} & 13.9451  & 0.3445  & 0.1463  & 0.0488  \\
    \multicolumn{2}{l}{GIFD-$f$} & 18.6457  & 0.2320  & 0.3916  & 0.0180  \\
    \multicolumn{2}{l}{GIFD-$e$} & 19.4662  & 0.1900  & 0.4383  & 0.0161  \\
    \multicolumn{2}{l}{GIFD} & \textbf{20.0534} & \textbf{0.1559} & \textbf{0.4713} & \textbf{0.0141} \\ \bottomrule
    
    \end{tabular}%
    }
    \vspace{0.5em}
    \caption{ImageNet}
    \end{subtable}

    \centering
    \begin{subtable}[t]{\linewidth}
    \centering
     \resizebox{0.8\linewidth}{!}{
     \begin{tabular}{lccccc} \toprule
    \multicolumn{2}{l}{\multirow{2}[1]{*}{Method}} & \multicolumn{4}{c}{Metric} \\ \cmidrule(lr){3-6}
     & & \multicolumn{1}{c}{PSNR$\uparrow$} & \multicolumn{1}{c}{LPIPS$\downarrow$} & \multicolumn{1}{c}{SSIM$\uparrow$} & \multicolumn{1}{c}{MSE$\downarrow$} \\  \midrule
    \multicolumn{2}{l}{GIFD-$z$} & 16.9947  & 0.1351  & 0.3931  & 0.0263  \\
    \multicolumn{2}{l}{GIFD-$f$} & 20.2506  & 0.1462  & 0.5210  & 0.0123  \\
    \multicolumn{2}{l}{GIFD-$e$} & 20.5839  & 0.1267  & 0.5412  & 0.0119  \\
    \multicolumn{2}{l}{GIFD} &  \textbf{21.3368} & \textbf{0.1023} & \textbf{0.5768} & \textbf{0.0098} \\ \bottomrule
    
    \end{tabular}%
    }
     \vspace{0.5em}
    \caption{FFHQ}
    \vspace{-1em}
    \end{subtable}

\label{ablation_table}%
\end{table}%

\textbf{Ablation Study on the Proposed Techniques.} We conduct ablation experiments on both datasets to further verify the effectiveness of each proposed technique. Specifically, we design three variants of GIFD. GIFD-$z$ only searches the latent space. GIFD-$f$ starts to search the intermediate feature domain without the $l_1$ ball limitation and outputs the final results from the last searched intermediate layer. Based on GIFD-$f$, GIFD-$e$ selects outputs from the layer with the least matching error. GIFD is GIFD-$e$ plus the $l_1$ ball limitation. The results of Table \ref{ablation_table} show that each aforementioned technique can further improve performance.

\begin{figure}[t]
	\centering
	\includegraphics[width=0.8\linewidth]{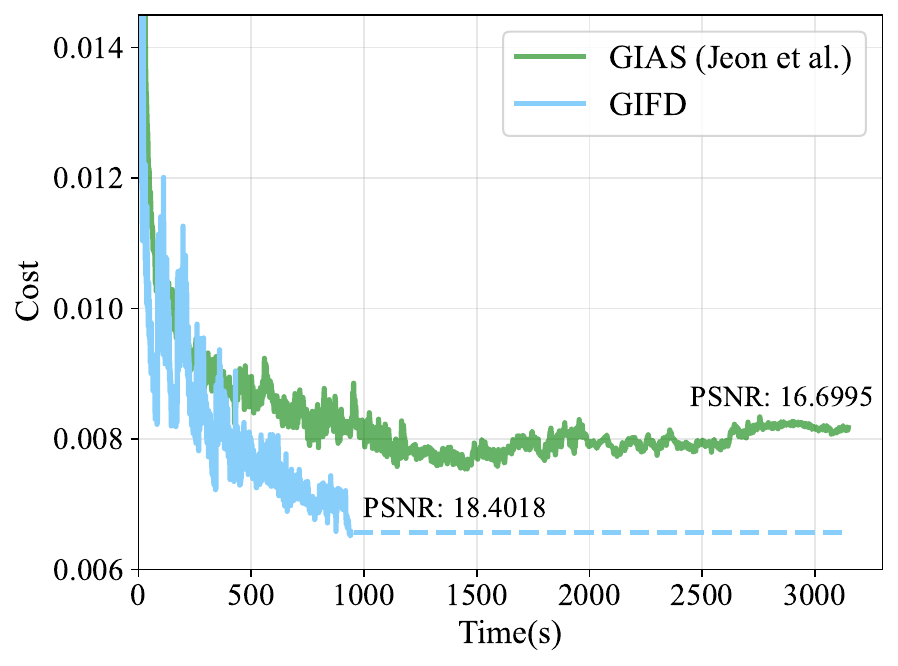}
	\centering
	\caption{The cost function over time of GIAS and GIFD with $B=4$. We run 4 trials and calculate the average values. Both methods execute a total of 8000 iterations.}
	\label{cost-function curve}
\end{figure}

\textbf{Reconstruction Time Analysis}. GIAS \cite{jeon2021gradient}, which alternates between searching the latent space and the parameter space of the generative model, generally yields the strongest performance among prior methods. A series of experiments has demonstrated that our method achieves consistent improvement over GIAS. Moreover, GIFD only searches the feature domain, whose optimized parameters are far fewer compared to the full generator parameters required by GIAS. Therefore, GIFD is expected to have an advantage in inference speed. In Figure \ref{cost-function curve}, we plot the cost-time curve for the intermediate feature searching phase of GIFD and the parameter space searching phase of GIAS. 

\begin{table}[t]
  \centering
 \scriptsize 
    \caption{Convergence time and PSNR results of GIFD and baselines that use the same distance metric at batch size 4.}
    \resizebox{7cm}{!}{\begin{tabular}{lccc} \toprule
    Method & IG \cite{geiping2020inverting}    & GIAS \cite{jeon2021gradient}  & \textbf{GIFD} \\     \midrule
    Time(s)↓ & \textbf{726.8008} & 1360.7808 & 907.727 \\
    PSNR↑  & 14.1896 & 16.6995 & \textbf{18.4018} \\
    Loss↓  & 0.008029 & 0.007803 & \textbf{0.006865} \\ \bottomrule
    \end{tabular}%
    }
  \label{timecost}%
\end{table}%

As expected, GIFD consumes less than one-third of the time required by GIAS while achieving a lower loss value. Moreover, GIAS requires repeatedly learning a dedicated generator for each reconstructed image, leading to substantial GPU memory consumption compared to the efficient GIFD.


\textbf{Computational Cost Scaling with Larger Image and Model Sizes}. To further investigate the computational costs under larger settings, we evaluate GIFD and GIAS under varying image resolutions and FL model sizes, and report the per-iteration time, GFLOPs, peak GPU memory, and energy in Tables \ref{tab:cost}. As expected, the computational cost of both attacks shows an overall upward trend as the resolution or the model size grows. Nevertheless, under every setting, GIFD consistently achieves higher PSNR than GIAS, while incurring a substantially lower computational cost, \eg, nearly halving per-iteration time and reducing peak GPU memory by about $50\%$ at $32\times32$. These demonstrate that GIFD achieves a favorable trade-off between reconstruction quality and computational cost across different image and model sizes.

\begin{table}[t]
  \centering
  \scriptsize
  \setlength{\tabcolsep}{3pt}

  \caption{Computational cost of GIAS and GIFD under different image resolutions and FL model sizes (per iteration). All experiments are conducted on ImageNet with $B=2$.}
  \label{tab:cost}

  \begin{subtable}{\linewidth}
    \centering
    \begin{tabular}{llcccc}
      \toprule
      \multirow{2}{*}{Metric} & \multirow{2}{*}{Method}
      & \multicolumn{4}{c}{Resolution} \\
      \cmidrule(lr){3-6}
      & & $32\times32$ & $64\times64$ & $128\times128$ & $224\times224$ \\
      \midrule
      \multirow{2}{*}{PSNR$\uparrow$}
      & GIAS \cite{jeon2021gradient}
      & 19.74 & 13.59 & 11.61 & 10.98 \\
      & \textbf{GIFD}
      & \textbf{20.35} & \textbf{14.62} & \textbf{12.37} & \textbf{11.71} \\
      \midrule
      \multirow{2}{*}{Time (s)$\downarrow$}
      & GIAS \cite{jeon2021gradient}
      & 0.27 & 0.28 & 0.34 & 0.48 \\
      & \textbf{GIFD}
      & \textbf{0.14} & \textbf{0.15} & \textbf{0.15} & \textbf{0.21} \\
      \midrule
      \multirow{2}{*}{GFLOPs$\downarrow$}
      & GIAS \cite{jeon2021gradient}
      & 510.35 & 540.46 & 660.94 & 1084.05 \\
      & \textbf{GIFD}
      & \textbf{442.19} & \textbf{482.31} & \textbf{642.77} & \textbf{992.24} \\
      \midrule
      \multirow{2}{*}{Peak mem. (MiB)$\downarrow$}
      & GIAS \cite{jeon2021gradient}
      & 3511.6 & 3527.5 & 3522.8 & 4145.2 \\
      & \textbf{GIFD}
      & \textbf{1763.5} & \textbf{1782.4} & \textbf{1877.5} & \textbf{2586.8} \\
      \midrule
      \multirow{2}{*}{Energy (J)}
      & GIAS \cite{jeon2021gradient}
      & 47.7 & 62.1 & 73.9 & 120.9 \\
      & \textbf{GIFD}
      & \textbf{26.4} & \textbf{29.0} & \textbf{34.6} & \textbf{53.5} \\
      \bottomrule
    \end{tabular}
    \vspace{1em}
    \caption{Image sizes}
    \label{tab:cost_resolution}
  \end{subtable}

  \vspace{6pt}

  \begin{subtable}{\linewidth}
    \centering
    \begin{tabular}{llcccc}
      \toprule
      \multirow{2}{*}{Metric} & \multirow{2}{*}{Method}
      & \multicolumn{4}{c}{FL Model} \\
      \cmidrule(lr){3-6}
      & & ResNet-18 & ResNet-34 & ResNet-50 & ResNet-101 \\
      \midrule
      \multirow{2}{*}{PSNR$\uparrow$}
      & GIAS \cite{jeon2021gradient}
      & 19.74 & 17.75 & 15.77 & 15.19 \\
      & \textbf{GIFD}
      & \textbf{20.35} & \textbf{18.58} & \textbf{16.83} & \textbf{16.06} \\
      \midrule
      \multirow{2}{*}{Time (s)$\downarrow$}
      & GIAS \cite{jeon2021gradient}
      & 0.27 & 0.33 & 0.55 & 0.73 \\
      & \textbf{GIFD}
      & \textbf{0.14} & \textbf{0.19} & \textbf{0.24} & \textbf{0.40} \\
      \midrule
      \multirow{2}{*}{GFLOPs$\downarrow$}
      & GIAS \cite{jeon2021gradient}
      & 510.35 & 521.54 & 524.50 & 547.02 \\
      & \textbf{GIFD}
      & \textbf{442.19} & \textbf{457.05} & \textbf{460.88} & \textbf{490.74} \\
      \midrule
      \multirow{2}{*}{Peak mem. (MiB)$\downarrow$}
      & GIAS \cite{jeon2021gradient}
      & 3511.6 & 3636.0 & 4045.9 & 4511.5 \\
      & \textbf{GIFD}
      & \textbf{1763.5} & \textbf{1919.4} & \textbf{1990.7} & \textbf{2399.8} \\
      \midrule
      \multirow{2}{*}{Energy (J)}
      & GIAS \cite{jeon2021gradient}
      & 47.7 & 54.9 & 84.5 & 106.3 \\
      & \textbf{GIFD}
      & \textbf{26.4} & \textbf{35.3} & \textbf{40.8} & \textbf{67.5} \\
      \bottomrule
    \end{tabular}
    \vspace{1em}
    \caption{Model sizes}
    \label{tab:cost_model}
  \end{subtable}

\end{table}

\begin{table}[t]
  \centering
  \caption{Comparison of different attack methods on $256\times256$ ImageNet images, where DenseNet-121 is adopted as the victim FL model, and the batch size is set to $B=8$.}
  \label{tab:high_resolution}
  \resizebox{0.9\linewidth}{!}{
    \begin{tabular}{lcccc}
    \toprule
     Method & PSNR$\uparrow$ & LPIPS$\downarrow$ & SSIM$\uparrow$ & MSE$\downarrow$ \\
    \midrule
     IG \cite{geiping2020inverting} & 14.5706 & 0.8751 & 0.2955 & 0.0349 \\
     GI \cite{yin2021see} & 14.4432 & 0.8824 & 0.2988 & 0.0359 \\
     GGL \cite{li2022auditing} & 14.1275 & 0.6755 & 0.2979 & 0.0387 \\
     GIAS \cite{jeon2021gradient} & 14.8944 & 0.6230 & 0.3258 & 0.0328 \\
     \textbf{GIFD} & \textbf{16.1348} & \textbf{0.6185} & \textbf{0.4148} & \textbf{0.0244} \\
    \bottomrule
    \end{tabular}
  }
\end{table}

\begin{table}[t]
\centering
\caption{Comparison with recent gradient inversion methods on ImageNet, including diffusion-based GGDM and Venom, theory-driven SPEAR, and architecture-aware ARES.}
\label{tab:more attacks}
\resizebox{0.9\linewidth}{!}{
\begin{tabular}{lcccc}
\toprule
 Method & PSNR$\uparrow$ & LPIPS$\downarrow$ & SSIM$\uparrow$ & MSE$\downarrow$ \\
\midrule
 SPEAR \cite{dimitrov2024spear} & 13.8619 & 0.6056 & 0.3085 & 0.0548 \\
 ARES \cite{gong2026ares} & 12.1564 & 0.7920 & 0.1353 & 0.0794 \\
 GGDM \cite{gu2024federated} & 17.4445 & 0.4589 & 0.4438 & 0.0267 \\
 Venom \cite{hu2026venom} & 18.2813 & 0.7405 & 0.2595 & 0.0230 \\
 GIAS \cite{jeon2021gradient} & 17.4923 & 0.2536 & 0.3381 & 0.0236 \\
 \textbf{GIFD} & \textbf{20.0534} & \textbf{0.1559} & \textbf{0.4713} & \textbf{0.0141} \\
\bottomrule
\end{tabular}
}
\end{table}

\textbf{Higher-Resolution Image Recovery}. To investigate the scalability of GIFD to larger image sizes, we additionally conduct experiments on 256$\times$256 images using DenseNet-121 as the victim model. Table \ref{tab:high_resolution} shows that GIFD still effectively reconstructs the private images and achieves strong attack performance under this higher-resolution setting.

\begin{table}[t]
  \centering
  \caption{Paired $t$-tests between GIFD and baselines. We conduct three runs with different random initializations. The $t$-statistic and the $p$-value are computed for each baseline against GIFD, and \checkmark~indicates statistical significance at $p<0.05$.}
  \begin{subtable}[t]{\linewidth}
    \centering
    \resizebox{0.95\linewidth}{!}{
    \begin{tabular}{lcccc} \toprule
     Method & PSNR (mean$\pm$std) & $t$-statistic & $p$-value & $p<0.05$ \\ \midrule
     IG \cite{geiping2020inverting} & 16.8349$\pm$1.0842 & 6.998 & 0.0198 & \checkmark \\
     GI \cite{yin2021see} & 16.4877$\pm$0.5629 & 27.369 & 0.0013 & \checkmark \\
     GGL \cite{li2022auditing} & 13.1534$\pm$0.2811 & 66.237 & 0.0002 & \checkmark \\
     GIAS \cite{jeon2021gradient} & 17.5997$\pm$0.7655 & 10.286 & 0.0093 & \checkmark \\
     \textbf{GIFD} & \textbf{20.1605$\pm$0.4296} & —— & —— & —— \\ \bottomrule
    \end{tabular}
    }
    \vspace{0.5em}
    \caption{ImageNet}
  \end{subtable}

  \begin{subtable}[t]{\linewidth}
    \centering
    \resizebox{0.95\linewidth}{!}{
    \begin{tabular}{lcccc} \toprule
     Method & PSNR (mean$\pm$std) & $t$-statistic & $p$-value & $p<0.05$ \\ \midrule
     IG \cite{geiping2020inverting} & 15.2508$\pm$0.8976 & 38.657 & 0.0007 & \checkmark \\
     GI \cite{yin2021see} & 15.0955$\pm$0.6750 & 94.669 & 0.0001 & \checkmark \\
     GGL \cite{li2022auditing} & 15.4197$\pm$0.2986 & 31.954 & 0.0010 & \checkmark \\
     GIAS \cite{jeon2021gradient} & 20.2403$\pm$0.6718 & 32.976 & 0.0009 & \checkmark \\
     \textbf{GIFD} & \textbf{21.7541$\pm$0.6412} & —— & —— & —— \\ \bottomrule
    \end{tabular}
    }
    \vspace{0.5em}
    \caption{FFHQ}
  \end{subtable}
  \label{tab:significance}
\end{table}

\textbf{Comparison with Recent Gradient Inversion Attacks.} To further evaluate performance against recent GLA, we first reproduce two powerful diffusion model-based methods, GGDM \cite{gu2024federated} and Venom \cite{hu2026venom}. Moreover, we also consider the theory-driven SPEAR \cite{dimitrov2024spear} and the architecture-aware attack ARES \cite{gong2026ares}. Note that the two methods rely on rigorous assumptions. SPEAR requires a fully connected first layer for low-rank decomposition, so its Theorems 3.1 and 3.2 do not apply to mainstream architectures. ARES requires modifying the victim model parameters, which does not hold in realistic FL settings. For fairness, we faithfully reproduce them following their implementations under the same settings considered.

As shown in Table \ref{tab:more attacks}, GIFD still consistently outperforms these recent baselines across different metrics, while GIAS falls behind Venom on PSNR and MSE, and GGDM on SSIM. This again demonstrates the superiority of the proposed hierarchical feature-domain optimization of our GIFD method.

\textbf{Statistical Robustness}. To assess the statistical significance of our method, we repeat the main experiments three times with different random initializations and report the mean and standard deviation. We further conduct paired $t$-tests between GIFD and the competing methods. As in Table \ref{tab:significance}, GIFD consistently achieves low variance and statistically significant improvements ($p<0.05$) across the evaluated settings compared to the previous SOTA GIAS, indicating that the observed performance gains are statistically significant.

%% file: conclusions.tex
\section{Conclusion and Future Work}
We propose GIFD, a powerful gradient inversion attack capable of generalizing to unseen OOD data scenarios. By exploiting the generative priors of pre-trained GANs, GIFD optimizes the hierarchical feature domain to generate stable and high-quality reconstruction results. Moreover, the proposed label mapping technique further enhances the attack on OOD scenarios of label inconsistency.
Extensive experiments on widely adopted datasets and two GAN architectures validate the effectiveness of GIFD across a range of realistic and challenging scenarios. To mitigate the proposed threat, a possible defense strategy is to inject adversarial noise into the gradients to obfuscate the inversion process. 

\textbf{Extension to Other Modalities.} This work focuses on image data, and the proposed GAN-based priors and visual feature representations are only applicable to the image modality. Nevertheless, modality-specific generative priors could provide useful auxiliary information for gradient attacks in text, audio, or other modalities. Future work could extend the feature-domain optimization strategy of GIFD to generative models for other modalities, \ie, exploring their intermediate layers to improve reconstruction performance.

\textbf{Theoretical Analysis}. While the proposed hierarchical optimization strategy is empirically effective, its theoretical properties remain not fully understood. In particular, formal convergence guarantees, recovery guarantees over the hierarchy of generator manifolds, and a rigorous characterization of privacy leakage are beyond the scope of this work. Future research could develop a theoretical framework for hierarchical feature optimization and analyze whether intermediate feature representations can provide stronger recovery guarantees in gradient inversion attacks than latent-space optimization alone.

We hope that this work can inspire new perspectives on gradient inversion attacks under more realistic FL settings and motivate further research in this direction. We also hope that our work can shed light on the development of stronger privacy-preserving mechanisms to enhance the security and robustness of FL systems.

\section*{Acknowledgements}
This work is supported in part by the National Natural Science Foundation of China under grant 62425201, 62301189, 62576122, 62571298, Guangdong Basic and Applied Basic Research Foundation under grant 2026A1515011139.
